# PO-VINS: An Efficient and Robust Pose-Only Visual-Inertial State Estimator With LiDAR Enhancement

Hailiang Tang, Tisheng Zhang, Liqiang Wang, Guan Wang, and Xiaoji Niu

*Abstract*—The pose adjustment (PA) with a pose-only visual representation has been proven equivalent to the bundle adjustment (BA), while significantly improving the computational efficiency. However, the pose-only solution has not yet been properly considered in a tightly-coupled visual-inertial state estimator (VISE) with a normal configuration for real-time navigation. In this study, we propose a tightly-coupled LiDAR-enhanced VISE, named PO-VINS, with a full pose-only form for visual and LiDAR-depth measurements. Based on the pose-only visual representation, we derive the analytical depth uncertainty, which is then employed for rejecting LiDAR depth outliers. Besides, we propose a multi-state constraint (MSC)-based LiDAR-depth measurement model with a pose-only form, to balance efficiency and robustness. The pose-only visual and LiDAR-depth measurements and the IMU-preintegration measurements are tightly integrated under the factor graph optimization framework to perform efficient and accurate state estimation. Exhaustive experimental results on private and public datasets indicate that the proposed PO-VINS yields improved or comparable accuracy to sate-of-the-art methods. Compared to the baseline method LE-VINS, the state-estimation efficiency of PO-VINS is improved by 33% and 56% on the laptop PC and the onboard ARM computer, respectively. Besides, PO-VINS yields higher accuracy and robustness than LE-VINS by employing the proposed uncertainty-based outlier-culling method and the MSC-based measurement model for LiDAR depth.

## NOMENCLATURE

| | |
|---|---|
| $\boldsymbol{p}^{\mathrm{p}}$ | The coordinate $(u, v)$ in the image pixel frame. |
| $\boldsymbol{p}^{\mathrm{u}}$ | The coordinate in the normalized camera frame (u-frame) with a unit depth. |
| $\boldsymbol{p}^{\mathrm{r}}$ | The LiDAR point in the LiDAR frame (range frame, r-frame). |
| $\mathbf{q}, \mathbf{R}, \boldsymbol{\phi}$ | The attitude quaternion, rotation matrix, and rotation vector. |
| $\otimes$ | The quaternion product. |
| $\mathrm{Log}, \mathrm{Exp}$ | The transformation between the quaternion and rotation vector. |
| $\boldsymbol{p}^{\mathrm{w}}_{\mathrm{wb}}, \mathbf{q}^{\mathrm{w}}_{\mathrm{b}}$ | The IMU pose (body frame, b-frame) w.r.t the world frame (w-frame). |
| $\boldsymbol{v}^{\mathrm{w}}_{\mathrm{wb}}$ | The IMU velocity in the world frame. |
| $\boldsymbol{b}_g, \boldsymbol{b}_a$ | The gyroscope and accelerometer biases. |
| $\boldsymbol{p}^{\mathrm{b}}_{\mathrm{bc}}, \mathbf{q}^{\mathrm{b}}_{\mathrm{c}}$ | The camera-IMU extrinsic parameters. |
| $\boldsymbol{p}^{\mathrm{r}}_{\mathrm{rc}}, \mathbf{q}^{\mathrm{r}}_{\mathrm{c}}$ | The camera-LiDAR extrinsic parameters. |
| $t_d$ | The time-delay parameter between the camera and the IMU data. |
| $\boldsymbol{X}, \boldsymbol{x}$ | The state vector. |

## I. INTRODUCTION

THE visual-inertial navigation system (VINS), due to its lower cost and smaller size, has been broadly used in mobile robots and autonomous vehicles [1]. A single camera cannot estimate the absolute scale of the ego-motion, resulting in the scale uncertainty of the monocular camera [2]. The ego-motion scale can be retrieved by utilizing the low-cost micro-electro-mechanical system (MEMS) inertial measurement unit (IMU). Meanwhile, the IMU errors, such as the biases and scale-factor errors [3], can also be estimated using the visual measurements from the camera. Hence, the tightly-coupled VINS can achieve reliable and accurate relative positioning, especially in global navigation satellite system (GNSS)-denied environments [4].

Typically, the visual features [5], [6] are detected and tracked frame by frame to obtain measurements, or the raw pixels [7] and image patches [8] are directly treated as the measurements. With the visual measurements from the front end and the IMU measurements, state estimation [9] can be performed to determine the pose and velocity. Conventionally, the filtering-based methods, including the extended Kalman filter (EKF) [10], [11] and iterated extended Kalman filter (IEKF) [12], are employed to solve the state-estimation problem in VINS due to lower computational costs. Particularly, the multi-state constraint Kalman filter (MSCKF)-based VINS [10], [11], [13], [14] has exhibited quite good performance. Factor graph optimization (FGO) [15] has been proven more accurate than filtering in VINS [16], [17]. This is because the nonlinear error of the visual measurement can be reduced notably by multiple iterations. Hence, the optimization-based VINS has attracted a lot of attention, including feature-based [18], [19], [20], [21], [22] and direct methods [7], [23], [24].

Though consuming many computational resources, the optimization-based VINS can still achieve superior real-time performance, such as VINS-Mono [19], especially with the rapid development of computer technology. Commonly, the bundle adjustment (BA) [2], [9] is adopted to estimate the pose and visual landmark parameters simultaneously. However, the

This research is partly funded by the Major Program (JD) of Hubei Province (2023BAA026) and the National Natural Science Foundation of China (No.42374034). (*Corresponding author: Tisheng Zhang.*)

Hailiang Tang, Liqiang Wang, and Guan Wang are with the GNSS Research Center, Wuhan University, Wuhan 430079, China (e-mail: thl@whu.edu.cn; wlq.a@whu.edu.cn; wanguan@whu.edu.cn).

Tisheng Zhang and Xiaoji Niu are with the GNSS Research Center, Wuhan University, Wuhan 430079, China, and also with the Hubei Luojia Laboratory, Wuhan 430079, China (e-mail: zts@whu.edu.cn; xjniu@whu.edu.cn).

high dimensional parameter space of visual landmarks significantly increases the computational complexity in solving the state-estimation problem in VINS. In contrast, the dimension of the pose parameters is much lower. Researchers have studied new methods to improve the efficiency of the BA, such as incremental BA [25], [26], fixed-lag smoother or sliding-window optimizer [19], [27], and preserving sparsity [28]. Nevertheless, too many computational resources have been used to estimate the parameter space of visual landmarks. In some edge devices, it is essential to decrease the computational costs, especially when minimal processor cores are available for positioning tasks.

Recently, the two-view imaging geometry has been proven equivalent to a pair of pose-only constraints decoupling camera poses from visual landmark parameters [29]. This work has been extended to the multiple-view imaging geometry [30]. The pose-only state estimation is conducted by adopting a pose adjustment (PA) rather than the BA. Besides, the visual landmark parameters can be analytically reconstructed from the obtained camera poses [30]. Specifically, the PA is conducted by explicitly expressing the visual landmark parameter with the pose and feature measurements. The pose-only solution has demonstrated that computational efficiency is significantly improved by 2-4 orders of magnitude for three-dimensional (3D) scene reconstruction [30]. PIPO-SLAM [31], a visual-inertial SLAM system based on ORB-SLAM3 [22], employs a local PA rather than the BA in the local mapping thread. The employed local PA yields significantly decreased computational and memory costs when more than 100 camera poses and 1000 landmarks are employed for local mapping optimization [31]. Note that the local PA in PIPO-SLAM [31] is a visual-only optimization, the same as in [30], rather than a tightly-coupled visual-inertial optimization. Besides, using 100 cameras and 1000 landmarks is not a normal configuration for real-time navigation.

Consequently, the pose-only solution has not yet been properly considered in a tightly-coupled visual-inertial state estimator (VISE) with a typical configuration for real-time navigation. Specifically, the impact of the pose-only solution on the accuracy and efficiency of the real-time VINS is still unknown. Meanwhile, light detection and ranging (LiDAR) can measure long-distance accurate depth and can be employed to enhance the VINS [32], [33], [34]. In LE-VINS [34], LiDAR has been proven to notably improve the robustness and accuracy of the VINS by constructing a LiDAR-depth measurement in the state estimator. Hence, the inverse-depth parameter [35] of the visual landmark can be accurately estimated with the LiDAR-depth measurement, and the pose accuracy can be improved [34]. However, the state parameters of the visual landmarks will be removed from the state vector in the pose-only solution. The pose-only LiDAR-depth measurement model should be further studied for accurate and efficient state estimation.

In this study, we aim to construct a pose-only VISE with LiDAR enhancement, named PO-VINS, to achieve an efficient and robust state estimation. With the pose-only visual representation, we present a pose-only visual-reprojection measurement model in the tightly-coupled VISE rather than using the visual-only PA in PIPO-SLAM [31]. Besides, we derive the analytical depth uncertainty for visual landmarks from the pose-only visual representation and propose an uncertainty-based outlier-culling algorithm for LiDAR depths. Moreover, we present a novel multi-state constraint (MSC)-based LiDAR-depth measurement model to balance the efficiency and robustness. Hence, the LiDAR enhancement can be seamlessly integrated into the pose-only VISE. The main contributions of this study can be listed as follows:

● A tightly-coupled visual-inertial state estimator with LiDAR enhancement is presented under the framework of FGO. The proposed state estimator is a complete pose-only solution for visual-reprojection and LiDAR-depth measurements, which balances efficiency and robustness.

● The analytical uncertainty for visual landmark depth is derived from the pose-only visual presentation. Hence, an uncertainty-based outlier rejection is proposed to detect and reject LiDAR-depth outliers, which has been proven to improve the robustness of the state estimation.

● The famous MSC method is adopted to build a pose-only LiDAR-depth measurement model by combining all visual and LiDAR-depth observations of a landmark into a single measurement. Hence, the measurement number can be decreased, and the estimation efficiency can be improved. Besides, it can also reduce the impact of the LiDAR depth outliers and improve the estimation robustness.

● Comprehensive experiments are conducted on public and private datasets with different carriers, including a low-speed robot, a handheld device, and a vehicle. We also verify the real-time performance of PO-VINS on an embedded onboard ARM computer.

The remainder of this paper is organized as follows. We first briefly discuss the related works about the VINS and the LiDAR-enhanced VINS. The system overview is described in Section III. Section IV presents the pose-only visual representation and derives the analytical depth uncertainty for landmarks. Then, the proposed LiDAR-enhanced VISE is given in Section V, with a full pose-only form for visual and LiDAR-depth measurements. The experiments and results are shown and discussed in Section VI for quantitative evaluation. Finally, we conclude the proposed PO-VINS.

## II. RELATED WORKS

In this study, we aim to construct a tightly-coupled VINS with LiDAR enhancement. Hence, the related works can be categorized into VISE and LiDAR-enhanced VISE. The LiDAR enhancement described here denotes the method using the LiDAR depth to improve the accuracy and robustness of the VISE. Tightly-coupled methods have been proven more accurate than loosely-coupled methods, and thus we will only discuss the former.

### A. Visual-Inertial State Estimator

In the early stage, the visual-inertial estimator is modeled using the Kalman filter [9] due to the limitations in computing resources. MonoSLAM is a real-time monocular simultaneous localization and mapping (SLAM) system that runs at 30 Hz with a standard personal computer within the EKF framework [36]. FEJ-EKF is proposed to improve the estimator's consistency by employing the first-estimates Jacobians (FEJ) [37]. The IEKF is also adopted for visual-inertial state estimation [12]. An MSCKF-based estimator is presented for vision-aided inertial navigation without including the visual landmarks in the state vector [10]. The delayed linearization in MSCKF avoids the computational burden and information loss, and thus, MSCKF yields improved precision and efficiency [10]. MSCKF 2.0 builds upon MSCKF by adopting the FEJ to achieve consistent state estimation and perform online estimation of the camera-IMU extrinsic parameters [11]. Due to superior real-time performance, the MSCKF has become a famous VISE framework. Hence, many recent works have focused on MSCKF-based estimators, such as R-VIO [13] and OpenVINS [14]. However, the one-time linearization for nonlinear visual measurements in filtering-based methods possibly introduces large linearization errors into the estimator and degrades performance [16].

In contrast, the nonlinear visual measurements can be repeatedly linearized by using the FGO [15]. Hence, the FGO should be more accurate than the filter for visual-inertial estimation [16], [17], though with higher computation complexity. The FGO or the nonlinear optimization is usually implemented by employing a sliding-window estimator [38] to reduce the dimensions of the state vector and improve estimation efficiency. Besides, the marginalization is conducted using the Schur complement [39] by converting all marginalized states into a prior to avoid loss of information [7], [27], [38]. OKVIS [27] is a keyframe-based visual-inertial odometry (VIO) using nonlinear optimization and marginalization, exhibiting improved accuracy than MSCKF-based methods. VINS-Mono [19] further employs a robust visual-inertial initialization and online relocalization using a global pose graph optimization (PGO). Kimera [20] includes a similar VIO module developed based on the famous nonlinear optimization library GTSAM [40]. ORB-SLAM-VI [18] and ORB-SLAM3 [22], which are both based on ORB-SLAM [41], [42], also employ the marginalization to construct the prior. BASALT [21] extracts relevant information from VIO using nonlinear factor recovery [43] for visual-inertial mapping. The direct spare method [7] is incorporated into VI-DOS [44], which includes dynamic marginalization. DM-VIO [24] is a delayed marginalization VIO that maintains a second factor graph for marginalization, yielding improved accuracy.

The above optimization-based methods employ marginalization to improve real-time performance without information loss. Incremental BA, such as ICE-BA [25] and iSAM2 [26], [40], can also enhance the efficiency of VISE by only updating limited states that are related to the observations. However, the incremental BA is mainly designed for global mapping rather than odometry. Except for these methods, using different feature parametrization, such as inverse depth [35], ParallaxBA [45], and PMBA [46], may result in various iterations and running time. Nevertheless, the high dimensional parameter space of visual landmarks is the main obstacle for efficient visual-inertial state estimation. The pose-only solution, *i.e.,* PA, has exhibited higher efficiency than BA for large-scale visual-only state estimation [30], [31]. Nevertheless, the pose-only solution has not yet been considered for a real-time tightly-coupled VISE, in which the dimension of landmark parameters is very limited. Specifically, the effects of the pose-only solution on the efficiency and accuracy of tightly-coupled VISE have not been demonstrated.

### B. LiDAR-Enhanced Visual-Inertial State Estimator

The LiDAR-enhanced VISE can be further classified into direct and feature-based methods [34]. Direct methods employ LiDAR depths for visual pixels, and the pose estimation is achieved by conducting photometric optimization. In feature-based methods, visual features are associated with LiDAR depths, and the BA is conducted to perform state estimation.

As the direct visual pixels are used in direct methods, sparse LiDARs, such as a 16-beam spinning LiDAR, can be adopted to provide depths. A direct laser-visual odometry is achieved by conducting a photometric-image alignment with occlusion handling [47]. DVL-SLAM is a similar direct visual SLAM system using window-based optimization to obtain an estimated pose, but it does not consider the occlusion. Accurate LiDAR depths have also been employed for pixels in tightly-coupled LiDAR-visual-inertial odometry (LVIO), such as R3LIVE [48] and FAST-LIVO [49]. However, the occlusion should be considered carefully to avoid wrong LiDAR depth usages, or it will degrade accuracy severely. Besides, the impact of the camera-LiDAR extrinsic errors is more notable in the direct methods, as the LiDAR depth at the corresponding pixel will be used.

Hence, many LiDAR-enhanced VISEs employ feature-based methods, in which visual features are first detected and then associated with LiDAR depths. In DEMO [32], visual features and LiDAR point clouds are associated in a unit sphere of the camera frame. This depth-association method is adopted by [50], [51], but the LiDAR depth is treated as constant without considering the depth error. LE-VINS [34] proposes a LiDAR-depth measurement model that constructs constraints to visual landmark states while considering the depth-association errors. In contrast, the depth association is achieved in the image plane in [33], [52], [53], resulting in high computational costs when projecting the point clouds into the image plane. A voxel-map-based depth-association method is used in [54], [55], but the accurate LiDAR depth is only treated as an initial value. A non-repetitive solid-state LiDAR is employed in CamVox [56] to build depth images, which are incorporated into ORB-SLAM2 [42] with RGB images to achieve an RGB-D SLAM. However, the non-overlapping area between the LiDAR and camera field of view (FOV) in CamVox will be partially wasted.

In conclusion, feature-based methods should be more suitable for LiDAR-enhanced VISE in terms of practicality and reliability. However, wrong or inaccurate depth associations may occur, no matter what depth-association methods are used. The reason is mainly because the occlusion may frequently happen on object edges, such as trees and vehicles. For example, a visual feature is detected on the edge of a tree, and it may associated with a LiDAR point on a wall behind the tree due to camera-LiDAR extrinsic errors. Hence, it is critical to reject LiDAR-depth outliers to improve the robustness. Besides, the LiDAR-depth measurement model should be further improved to perform the high-efficiency characteristic of the pose-only framework. Consequently, we can achieve efficient and robust LiDAR-enhanced VISE by constructing pose-only visual and LiDAR-depth measurements.

## III. System Overview

The proposed PO-VINS is built upon our previous work LE-VINS [34] by further incorporating the pose-only solution. Specifically, we derive the analytical uncertainty of the landmark depth from the pose-only representation and employ the uncertainty to reject LiDAR depth outliers. Moreover, we propose pose-only visual-reprojection and MSC-based LiDAR-depth measurements within the tightly-coupled framework.

The system pipeline of the proposed PO-VINS is depicted in Fig. 1. We employ an inertial navigation system (INS)-centric [4] processing framework to perform the short-time accuracy of the INS. The INS is initialized by gravity leveling [3] to estimate roll and pitch angles roughly while the initial position, velocity, and heading angle are set to zero. We can also derive the initial gyroscope biases during stationary conditions. Note that in-motion visual alignment methods [19] can also be adopted, but they are out of the scope of this paper. The high-frequency INS mechanization can be conducted with the initialized states, and the obtained INS pose will be used for the following camera and LiDAR data processing.

With the continuous visual image frames derived from the camera, the *Shi-Tomasi* [5] feature points are detected parallelly in separated image grids to improve efficiency. The prior INS pose is then employed to assist the Lukas-Kanade optical flow [57] to enhance the tracking continuity. The tracking efficiency can also be improved with the INS aiding by reducing the usage of the image pyramid [4]. The visual keyframe is also selected by judging the keyframe interval and relative motions [4]. Only the feature points in the keyframes will be added to the landmark manager for state estimation to perform the short-term accuracy of the INS and improve efficiency.

Meanwhile, the accumulated LiDAR point clouds are projected to the visual keyframe time using the high-frequency INS pose with distortion removed. Hence, the feature pairs and LiDAR points can be associated in a unit sphere of the camera frame, and LiDAR depths for corresponding landmarks can be estimated by employing a plane-fitting method [34]. Several outlier-culling methods have been adopted to remove LiDAR depth outliers, such as using only foreground point clouds and plane checking [34]. However, LiDAR depth outliers may exist, especially in unstructured environments. Thus, we derive the analytical depth uncertainty for visual landmarks from the pose-only representation and employ the uncertainty to remove LiDAR depth outliers.

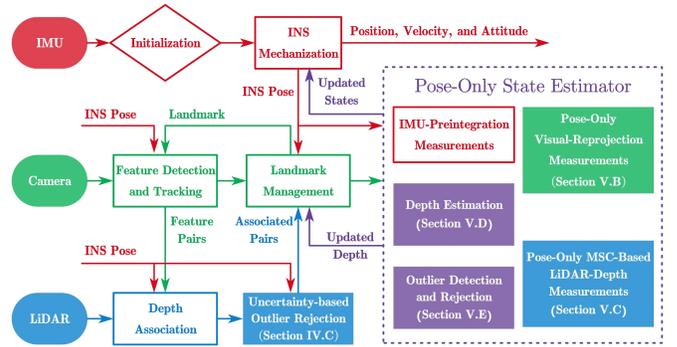

Fig. 1. System overview of the proposed PO-VINS.

Finally, we obtain a series of visual features, some associated with LiDAR depths. The visual features, LiDAR depths, and IMU measurements are tightly integrated using sliding-window-based FGO to perform maximum-a-posterior estimation. The three-axis gyroscope and accelerometer observations from the IMU are used to construct an IMU-preintegration measurement model [58]. Those visual features without LiDAR depth are utilized to build pose-only visual-reprojection measurements (VRM). Hence, the landmark states are not incorporated into the state vector; thus, the dimension of the state vector is reduced. The proposed VRM is in a tightly-coupled form, as the IMU pose states are adopted in the state vector. For those visual features with LiDAR depths, we present MSC-based LiDAR-depth measurements (LDM) in pose-only form to compress the measurement number, thus improving efficiency. The system is also more robust to LiDAR-depth outliers by combining all feature observations of a landmark and LiDAR depth into one measurement. Pose-only state estimation is achieved by integrated visual-reprojection, LiDAR-depth, and IMU-preintegration measurements to conduct nonlinear optimization. After that, the depths for visual landmarks are updated, and visual feature outliers are detected and rejected with the updated depths. Meanwhile, the updated INS states, including navigation states and IMU biases, are returned to the INS mechanization module for real-time high-frequency navigation.

## IV. Pose-Only Representation for Visual Landmarks

This section presents the methodology of the pose-only visual representation. We first derive the pose-only representation for visual landmarks from the multiple-view geometry. The analytical depth uncertainty of the landmark can be obtained by conducting error propagation. Based on the derived uncertainty, the LiDAR depth outliers can be further detected and rejected.

### A. Pose-Only Visual Representation

Consider a 3D visual landmark in the world frame (w-frame) $\boldsymbol{p}^w$ observed in several images, as depicted in Fig. 2. The origin of the camera frame (c-frame) is represented by $\boldsymbol{O}_c$.

$\boldsymbol{p}^{\mathrm{p}} = (u, v)$ is the observed feature in the image plane, *i.e.,* pixel frame (p-frame). $\boldsymbol{p}^{\mathrm{p}}$ can be converted into the normalized camera frame (u-frame) using the back projection function $\pi_c^{-1}$ and camera intrinsic parameters [2], denoted as $\boldsymbol{p}^{\mathrm{u}} = \left(x^{\mathrm{u}}, y^{\mathrm{u}}, 1\right)$. The projection equation of the 3D visual landmark $\boldsymbol{p}^{\mathrm{w}}$ in the u-frame can be written as

$$\boldsymbol{p}^{\mathrm{u}} = \frac{1}{z^{\mathrm{c}}} \boldsymbol{p}^{\mathrm{c}} = \frac{1}{z^{\mathrm{c}}} \left(\mathbf{R}_c^{\mathrm{w}}\right)^{-1} \left(\boldsymbol{p}^{\mathrm{w}} - \boldsymbol{p}_{\mathrm{wc}}^{\mathrm{w}}\right), \quad (1)$$

where $\boldsymbol{p}^{\mathrm{c}} = \left(x^{\mathrm{c}}, y^{\mathrm{c}}, z^{\mathrm{c}}\right)$ is the coordinate of the visual landmark in the c-frame; $\left\{\boldsymbol{p}_{\mathrm{wc}}^{\mathrm{w}}, \mathbf{R}_c^{\mathrm{w}}\right\}$ is the camera pose w.r.t the w-frame.

For two images $\mathbf{F}_\varsigma$ and $\mathbf{F}_\eta$ in Fig. 2, we can derive the following equation from (1) as

$$z^{c_\eta} \boldsymbol{p}^{\mathrm{u}_\eta} = z^{c_\varsigma} \mathbf{R}_{c_\varsigma}^{c_\eta} \boldsymbol{p}^{\mathrm{u}_\varsigma} + \boldsymbol{p}_{c_\eta c_\varsigma}^{c_\eta}, \quad (2)$$

where $\left\{\boldsymbol{p}_{c_\eta c_\varsigma}^{c_\eta}, \mathbf{R}_{c_\varsigma}^{c_\eta}\right\}$ is the relative transformation from the c-frame $c_\varsigma$ to the c-frame $c_\eta$ and they can be written as

$$\begin{cases} \boldsymbol{p}_{c_\eta c_\varsigma}^{c_\eta} = \left(\mathbf{R}_{c_\eta}^{\mathrm{w}}\right)^{-1} \left(\boldsymbol{p}_{\mathrm{wc}_\varsigma}^{\mathrm{w}} - \boldsymbol{p}_{\mathrm{wc}_\eta}^{\mathrm{w}}\right) \\ \mathbf{R}_{c_\varsigma}^{c_\eta} = \left(\mathbf{R}_{c_\eta}^{\mathrm{w}}\right)^{-1} \mathbf{R}_{c_\varsigma}^{\mathrm{w}} \end{cases}. \quad (3)$$

Left multiply the skew-symmetric matrix $\boldsymbol{p}^{\mathrm{u}_\eta}{}_\times$ [59] on both sides of (2), and we can obtain

$$z^{c_\varsigma} \ \boldsymbol{p}^{\mathrm{u}_\eta}{}_\times \mathbf{R}_{c_\varsigma}^{c_\eta} \boldsymbol{p}^{\mathrm{u}_\varsigma} = -\boldsymbol{p}^{\mathrm{u}_\eta}{}_\times \boldsymbol{p}_{c_\eta c_\varsigma}^{c_\eta}. \quad (4)$$

Taking the magnitude in (4), the landmark depth $z^{c_\varsigma}$ in the c-frame $c_\varsigma$ can be written as

$$z^{c_\varsigma} = \frac{\left\| \boldsymbol{p}^{\mathrm{u}_\eta}{}_\times \boldsymbol{p}_{c_\eta c_\varsigma}^{c_\eta} \right\|}{\theta_{\varsigma, \eta}} \triangleq d_\varsigma^{(\varsigma, \eta)}, \quad (5)$$

where $\theta_{\varsigma, \eta} = \left\| \boldsymbol{p}^{\mathrm{u}_\eta}{}_\times \mathbf{R}_{c_\varsigma}^{c_\eta} \boldsymbol{p}^{\mathrm{u}_\varsigma} \right\|$. Similarly, we can obtain $z^{c_\eta}$ by left multiply $\mathbf{R}_{c_\varsigma}^{c_\eta} \boldsymbol{p}^{\mathrm{u}_\varsigma}{}_\times$ on both sides of (2) as

$$z^{c_\eta} = \frac{\left\| \mathbf{R}_{c_\varsigma}^{c_\eta} \boldsymbol{p}^{\mathrm{u}_\varsigma}{}_\times \boldsymbol{p}_{c_\eta c_\varsigma}^{c_\eta} \right\|}{\theta_{\varsigma, \eta}} \triangleq d_\eta^{(\varsigma, \eta)}. \quad (6)$$

Hence, the pose-only constraint for two-view imaging geometry in (2) can be written as

$$d_\eta^{(\varsigma, \eta)} \boldsymbol{p}^{\mathrm{u}_\eta} = d_\varsigma^{(\varsigma, \eta)} \mathbf{R}_{c_\varsigma}^{c_\eta} \boldsymbol{p}^{\mathrm{u}_\varsigma} + \boldsymbol{p}_{c_\eta c_\varsigma}^{c_\eta}. \quad (7)$$

The pose-only constraint in (7) has been proven to be equivalent to the two-view imaging geometry [29]. Supposing that $\mathbf{F}_\varsigma$ and $\mathbf{F}_\eta$ are two anchored frames of the visual landmark $\boldsymbol{p}^{\mathrm{w}}$, we can derive a set of constraints $C(\varsigma, \eta)$ as

$$C(\varsigma, \eta) = \left\{ d_j^{(\varsigma, j)} \boldsymbol{p}^{\mathrm{u}_j} = d_\varsigma^{(\varsigma, \eta)} \mathbf{R}_{c_\varsigma}^{c_j} \boldsymbol{p}^{\mathrm{u}_\varsigma} + \boldsymbol{p}_{c_j c_\varsigma}^{c_j}, j \neq \varsigma \right\}, \quad (8)$$

where $\mathbf{F}_j$ denotes another observed frame, as shown in Fig. 2. The equation (8) is the pose-only visual representation for multiply-view geometry and has been proved equivalent to the projection equation (1) [30]. The key idea in pose-only visual representation (8) is that we can explicitly use the pose of the anchored frames and feature observations to express the landmark state, i.e., the depth. Consequently, the visual landmark states can be removed from the state vector, and thus the state-estimation efficiency can be improved.

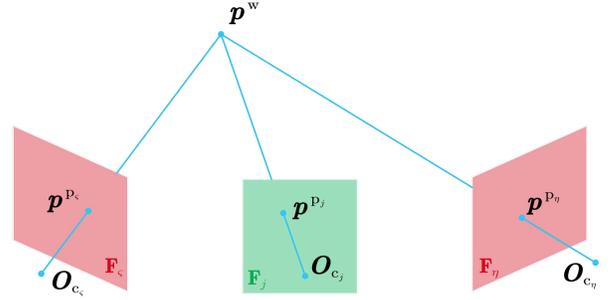

Fig. 2. An illustration of the multi-view geometry. The red frames $\mathbf{F}_\varsigma$ and $\mathbf{F}_\eta$ represent the anchored frames of the landmark $\boldsymbol{p}^{\mathrm{w}}$. Note that only keyframes are employed for state estimation.

Consequently, the pose-only visual measurements can be derived from (8). It should be noted that the anchored frames $\mathbf{F}_\varsigma$ and $\mathbf{F}_\eta$ should construct the largest parallax of the landmark $\boldsymbol{p}^{\mathrm{w}}$. Fortunately, the parameter $\theta_{\varsigma, \eta}$ can be employed to represent the magnitude of the parallax [30]. In practice, the anchored frame $\mathbf{F}_\varsigma$ is set to the first observed frame or the frame associated with LiDAR depth in PO-VINS to facilitate the sliding-window marginalization. The anchored frame $\mathbf{F}_\eta$ can be searched within the other frames to meet the largest parallax, thus maximizing visual constraints and improving accuracy.

### B. Analytical Depth Uncertainty for Visual Landmark

With the pose-only depth formula (5), the landmark depth $d_\varsigma^{(\varsigma, \eta)}$ in the anchored frame $\mathbf{F}_\varsigma$ can be analytically obtained using camera pose and feature observations. We can also derive the analytical depth uncertainty using error perturbation [3]. In tightly-coupled VISE, the camera pose is calculated by IMU pose states and camera-IMU extrinsic parameters. Due to the high short-term accuracy of the INS pose, the relative pose errors of the camera can be ignored within the sliding window, whose length is usually about several seconds. Hence, the depth uncertainty is mainly caused by feature noise. Typically, the feature noise $\sigma_p$ is manually set to a fixed value, such as in [14], [19], and it is set as $\sigma_p = 1.5$ pixels in PO-VINS. We can obtain the covariance matrix of feature points $\boldsymbol{\Sigma}_p^{\mathrm{u}}$ in the u-frame as

$$\boldsymbol{\Sigma}_p^{\mathrm{u}} = \begin{vmatrix} \left| \sigma_p / f_x \right|^2 & 0 \\ 0 & \left| \sigma_p / f_y \right|^2 \end{vmatrix}, \quad (9)$$

where $f_x$ and $f_y$ are focal lengths in intrinsic parameters.

According to the pose-only depth formulation (5), we can

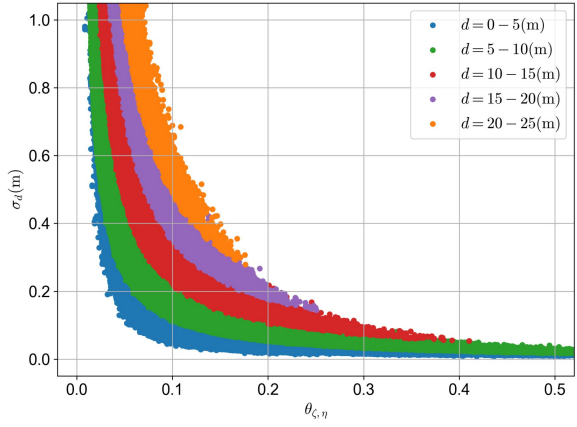

Fig. 3. The relationship between the depth STD $\sigma_d$ and the parallax parameter $\theta_{\varsigma,\eta}$. The results are obtained from real-world datasets.

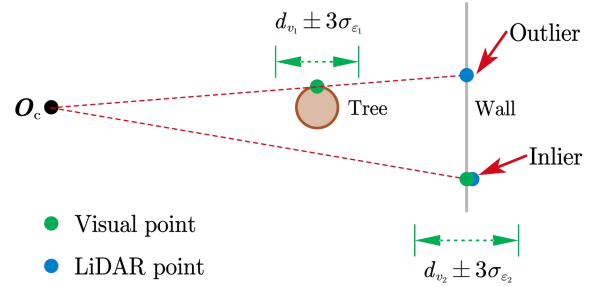

Fig. 4. An illustration of the LiDAR-depth outlier and inlier. The visual points denote the observed position of landmarks, where the LiDAR points represent the calculated position using associated LiDAR depths.

write the Jacobians w.r.t the feature error $\boldsymbol{p}^{u_\varsigma}$ as

$$\frac{\partial d_\varsigma^{(\varsigma,\eta)}}{\partial \boldsymbol{p}^{u_\varsigma}} = \left| -\frac{\beta}{\theta^3} \right| \left( \boldsymbol{p}^{u_\eta} \quad \mathbf{R}_{c_\varsigma}^{c_\eta} \right)^T \boldsymbol{p}^{u_\eta} \quad \mathbf{R}_{c_\varsigma}^{c_\eta} \boldsymbol{p}^{u_\varsigma} \Big|^T_{[0:1]}, \quad (10)$$

where the real number $\beta$ is expressed as

$$\beta = \left\| \boldsymbol{p}^{u_\eta} \times \boldsymbol{p}^{c_\eta}_{c_\eta c_\varsigma} \right\|. \quad (11)$$

Note the third dimension in (10) is not used, as the raw feature point is in two dimensions. Similarly, the Jacobians w.r.t the feature error $\boldsymbol{p}^{u_\eta}$ can be written as

$$\frac{\partial d_\varsigma^{(\varsigma,\eta)}}{\partial \boldsymbol{p}^{u_\eta}} = \left( -\frac{1}{\beta\theta} \left| \left( \boldsymbol{p}^{c_\eta}_{c_\eta c_\varsigma \times} \right)^T \boldsymbol{p}^{u_\eta} \times \boldsymbol{p}^{c_\eta}_{c_\eta c_\varsigma} \right| \right. \\ \left. + \frac{\beta}{\theta^3} \left| \left( \mathbf{R}_{c_\varsigma}^{c_\eta} \boldsymbol{p}^{u_\varsigma} \right)^T \boldsymbol{p}^{u_\eta} \quad \mathbf{R}_{c_\varsigma}^{c_\eta} \boldsymbol{p}^{u_\varsigma} \right| \right)_{[0:1]}. \quad (12)$$

Finally, the uncertainty $\boldsymbol{\Sigma}_d = \sigma_d^2$ of the landmark depth $d_\varsigma^{(\varsigma,\eta)}$ can be analytically calculated as

$$\boldsymbol{\Sigma}_d = \frac{d_\varsigma^{(\varsigma,\eta)}}{\boldsymbol{p}^{u_\varsigma}}^T \boldsymbol{\Sigma}_p^u \frac{d_\varsigma^{(\varsigma,\eta)}}{\boldsymbol{p}^{u_\varsigma}} + \frac{d_\varsigma^{(\varsigma,\eta)}}{\boldsymbol{p}^{u_\eta}}^T \boldsymbol{\Sigma}_p^u \frac{d_\varsigma^{(\varsigma,\eta)}}{\boldsymbol{p}^{u_\eta}}. \quad (13)$$

Fig. 3 shows the depth uncertainty $\sigma_d$, i.e., the standard deviation (STD), w.r.t different parallax parameters $\theta_{\varsigma,\eta}$ and depth. It exhibits that the depth uncertainty is minor when the parallax parameter is larger. Besides, if the depth is larger, the depth uncertainty is larger. The results satisfy the theory of the two-view geometry [2]. Hence, it demonstrates that the derived analytical depth uncertainty in (13) is correct. We also notice that the depth uncertainty $\sigma_d$ may be larger than several decimeters, which is far larger than the LiDAR measurement error. This is why the estimation accuracy can be improved by incorporating the LiDAR enhancement.

### C. Uncertainty-Based Outlier Rejection for LiDAR Depths

The depth uncertainty can be analytically calculated within the pose-only framework using (13). The depth uncertainty can be further employed to detect and reject LiDAR-depth outliers. Considering a visual landmark, its depth $d_v$ in the anchored frame $\mathbf{F}_\varsigma$ can be calculated using the camera pose of the two anchored frames $\mathbf{F}_\varsigma$ and $\mathbf{F}_\eta$ as in (5). Meanwhile, the depth uncertainty $\boldsymbol{\Sigma}_{d_v} = \sigma_{d_v}^2$ can be calculated using (13). Suppose the landmark is associated with a LiDAR depth $d_r$, and the LiDAR-depth has an STD $\sigma_{d_r}$. Here, the STD $\sigma_{d_r}$ is set to a fixed value of 0.1 m, which is determined by the depth-association methods in LE-VINS [34]. Note the depth $d_v$ and $d_r$ are all in the c-frame.

The proposed uncertainty-based outlier rejection for LiDAR depths is based on 3-sigma verification. The difference between the calculated depth $d_v$ and LiDAR depth $d_r$ is written as

$$\varepsilon_d = d_v - d_r. \quad (14)$$

According to the property of random variables, the STD of the difference $\varepsilon_d$ can be calculated as

$$\sigma_{\varepsilon_d} = \sqrt{\sigma_{d_v}^2 + \sigma_{d_r}^2}. \quad (15)$$

Hence, if the difference $\varepsilon_d$ is not within $\pm 3\sigma_{\varepsilon_d}$, the associated LiDAR depth should be treated as an outlier and rejected. It should be noted that the LiDAR-depth outliers will be removed. Nevertheless, the visual feature pairs associated with the LiDAR-depth outliers will be reserved for constructing visual measurements.

Fig. 4 exhibits an illustration of the outlier and inlier of LiDAR depths. As shown in Fig. 4, the outlier happens on a tree edge, and the visual point is on the tree while the LiDAR point is on the wall. The visual and LiDAR points are all on the wall for the inlier. According to the results, the LiDAR-depth outliers mainly occur in unstructured environments, especially with many foreground and background objects. We also show a qualitative result in Fig. 5. With the proposed outlier rejection method, fewer visual landmarks are associated with LiDAR depths (the green rectangle) in the unstructured leaves (within the big red rectangle). Meanwhile, it almost exhibits the same results on the ground and the wall, with or without using the proposed outlier-culling method, as shown in Fig. 5. All in all, the LiDAR-depth outliers can be detected and removed by using the presented uncertainty-based outlier rejection, and thus, the system robustness can be improved.

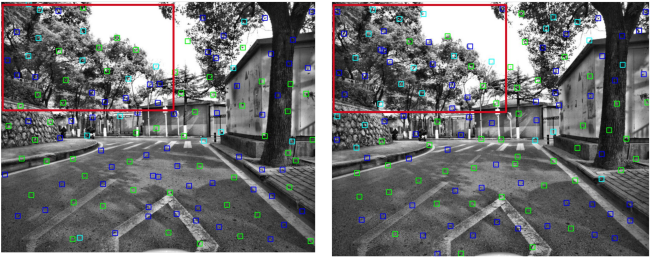

(a) Without outlier culling.    (b) With outlier culling.

Fig. 5. The impact of the outlier-culling method for LiDAR depths. The blue rectangles denote the uninitialized landmarks. The cyan and green rectangles denote the initialized landmarks, and the green rectangles are associated with LiDAR depth. The big red rectangles represent the unstructured areas, i.e., the tree leaves, where outliers may frequently occur. With the proposed outlier-culling algorithm, fewer landmarks are associated with LiDAR depth within the red rectangles.

## V. Pose-Only Visual-Inertial State Estimator with LiDAR Enhancement

This section presents the proposed pose-only VISE with LiDAR-depth enhancement. The state-estimation problem is first presented to discuss the FGO framework of PO-VINS, as shown in Fig. 6. Then, the visual-reprojection (VRM) and LiDAR-depth (LRM) measurements are all modeled in pose-only form. We also study the depth-estimation method, as the landmark depth will be employed for subsequent outlier culling and feature tracking. Finally, we briefly introduce the outlier-culling method for visual features.

### A. Tightly-Coupled State Estimator

The proposed tightly-coupled state estimator is based on the sliding-window optimizer, and the state vector $\boldsymbol{X}$ in PO-VINS can be defined as follow

$$
\begin{aligned}
\boldsymbol{X} &= \boldsymbol{x}_0, \boldsymbol{x}_1, ..., \boldsymbol{x}_n, \boldsymbol{x}_{\mathrm{c}}^{\mathrm{b}}, \\
\boldsymbol{x}_k &= \boldsymbol{p}_{\mathrm{wb}_k}^{\mathrm{w}}, \mathbf{q}_{\mathrm{b}_k}^{\mathrm{w}}, \boldsymbol{v}_{\mathrm{wb}_k}^{\mathrm{w}}, \boldsymbol{b}_{g_k}, \boldsymbol{b}_{a_k}, t_{d_k}, k \in [0, n], \\
\boldsymbol{x}_{\mathrm{c}}^{\mathrm{b}} &= \boldsymbol{p}_{\mathrm{bc}}^{\mathrm{b}}, \mathbf{q}_{\mathrm{c}}^{\mathrm{b}},
\end{aligned} \tag{16}
$$

where $\boldsymbol{x}_k$ is the IMU state at each time node, including the position $\boldsymbol{p}_{\mathrm{wb}}^{\mathrm{w}}$, the attitude quaternion $\mathbf{q}_{\mathrm{wb}}^{\mathrm{w}}$, and the velocity $\boldsymbol{v}_{\mathrm{wb}}^{\mathrm{w}}$ in the w-frame; $\boldsymbol{b}_g$ and $\boldsymbol{b}_a$ are the gyroscope and the accelerometer biases, respectively; b denotes the IMU body frame (b-frame); $n$ is the number of the IMU preintegration in the sliding window; $\boldsymbol{x}_{\mathrm{c}}^{\mathrm{b}}$ is the camera-IMU extrinsic parameters. The attitude quaternion $\mathbf{q}$ and the rotation matrix $\mathbf{R}$ are equivalent [59]. Note the time-delay parameter $t_d$ between the camera and IMU is modeled as a random walk [3] for better robustness, as the camera and IMU may not be well synchronized in some datasets. For convenience, the time-delay parameters for camera-IMU are omitted in the following parts. For details about the implementation of the time-delay model, we can refer to the open-sourced LE-VINS [34] on GitHub[1].

Note that the visual landmark state parameter, such as the inverse depth or the 3D position, is not contained in (16). Thus,

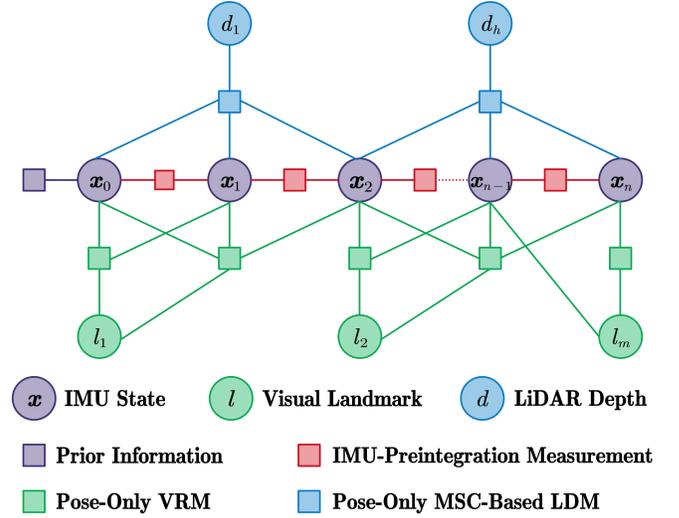

Fig. 6. The FGO framework of the proposed PO-VINS. Here, the VRM and LDM represent the visual-reprojection measurement and LiDAR-depth measurement, respectively. The pose-only VRM may include two or three states. The pose-only MSC-based LDM compresses all observations of a single landmark and the LiDAR depth in to one measurement. Note that the IMU velocity and biases are not depicted explicitly in the figure for better visualization.

the proposed state estimator is in pose-only form regarding the visual-reprojection and LiDAR-depth measurements. Besides, the proposed pose-only estimator is a tightly-coupled form. The state estimation in PO-VINS is achieved by constructing a nonlinear optimization problem, which can be solved by minimizing the sum of the Mahalanobis norm [9] of all measurements and the prior as

$$
\begin{aligned}
\arg\min_{\boldsymbol{X}} \frac{1}{2} & \sum_{\substack{j, \eta \tilde{,}l_i \\ i \in [1,m]}} \left\| \boldsymbol{e}^V\left(\tilde{\boldsymbol{z}}_{j,l_i}^{V,\eta}, \boldsymbol{X}\right) \right\|_{\boldsymbol{\Sigma}_{l_i}^{V,\eta}}^2 + \sum_{i \, [1,k]} \left\| \boldsymbol{e}^D\left(\tilde{\boldsymbol{z}}^{D_i}, \boldsymbol{X}\right) \right\|_{\boldsymbol{\Sigma}^{D_i}}^2 \\
& + \sum_{k \in [1,n]} \left\| \boldsymbol{e}^{Pre}\left(\tilde{\boldsymbol{z}}_{k-1,k}^{Pre}, \boldsymbol{X}\right) \right\|_{\boldsymbol{\Sigma}_{k-1,k}^{Pre}}^2 + \left\| \boldsymbol{e}^{Prior}\left(\tilde{\boldsymbol{z}}^{Prior}, \boldsymbol{X}\right) \right\|^2
\end{aligned} \tag{17}
$$

where $\boldsymbol{e}^V$ are residuals for the pose-only VRM; $l_i$ is a visual landmark, and $\mathbf{F}_\varsigma$ and $\mathbf{F}_\eta$ is its two anchored keyframes, mentioned in Section IV.A; $\mathbf{F}_j, j \neq \varsigma$ is another keyframe of the landmark $l_i$; $\boldsymbol{e}^D$ are residuals for the pose-only MSC-based LDM, which compresses all feature observations of a visual landmark and the associated LiDAR depth into one measurement; $\boldsymbol{e}^{Pre}$ are residuals for IMU-preintegration measurements; $\boldsymbol{e}^{Prior}$ are residuals for prior constraints, which are derived from the marginalization [7], [27], [38].

Ceres Solver [60], an open-sourced library for modeling and solving large optimization problems, is adopted in PO-VINS. Specifically, the Levenberg-Marquardt (LM) [15] algorithm is employed to solve the nonlinear optimization problem in (17). Meanwhile, we also adopt the Huber robust cost function [60] for VRM and LDM to reduce the effects of the visual-feature and LiDAR-depth outliers. Fig.6 depicts the FGO framework of the proposed PO-VINS.

The IMU-preintegration measurement constructs relative pose constraints between two consecutive IMU states while incorporating the velocity and IMU biases constraints. We follow our previous work to model the IMU-preintegration measurement [58], and the residuals can be written as

$$
\begin{aligned}
& \boldsymbol{e}^{Pre}\left(\tilde{\boldsymbol{z}}_{k-1,k}^{Pre}, \boldsymbol{X}\right) = \\
& \begin{bmatrix}
\left(\mathbf{R}_{b_{k-1}}^{w}\right)^{T}\begin{bmatrix}\boldsymbol{p}_{wb_{k}}^{w} - \boldsymbol{p}_{wb_{k-1}}^{w} - \boldsymbol{v}_{wb_{k-1}}^{w}\Delta t_{k-1,k} \\ -\frac{1}{2}\boldsymbol{g}^{w}\Delta t_{k-1,k}^{2}\end{bmatrix} - \Delta\tilde{\boldsymbol{p}}_{k-1,k}^{Pre} \\
\left(\mathbf{R}_{b_{k-1}}^{w}\right)^{T}\left(\boldsymbol{v}_{wb_{k}}^{w} - \boldsymbol{v}_{wb_{k-1}}^{w} - \boldsymbol{g}^{w}\Delta t_{k-1,k}\right) - \Delta\tilde{\boldsymbol{v}}_{k-1,k}^{Pre} \\
\mathrm{Log}\left(\left(\mathbf{q}_{b_{k}}^{w}\right)^{-1}\quad \mathbf{q}_{b_{k-1}}^{w}\quad \tilde{\mathbf{q}}_{k-1,k}^{Pre}\right) \\
\boldsymbol{b}_{g_{k}} - \boldsymbol{b}_{g_{k-1}} \\
\boldsymbol{b}_{a_{k}} - \boldsymbol{b}_{a_{k-1}}
\end{bmatrix},
\end{aligned}
\tag{18}
$$

where $\Delta\tilde{\boldsymbol{p}}_{k-1,k}^{Pre}$, $\Delta\tilde{\boldsymbol{v}}_{k-1,k}^{Pre}$, and $\tilde{\mathbf{q}}_{k-1,k}^{Pre}$ are the position, velocity, and attitude preintegration observations [58], respectively; $\mathbf{R}_{b}^{w}$ denotes the rotation matrix of the quaternion $\mathbf{q}_{b}^{w}$; $\boldsymbol{g}^{w}$ denotes the gravity vector in the w-frame; $\mathrm{Log}(\bullet)$ represents the transformation from quaternion to rotation vector [59]; $\Delta t_{k-1,k}$ is the time length between the two IMU states, i.e., the interval of visual keyframes. The covariance matrix $\boldsymbol{\Sigma}^{Pre}$ is obtained by noise propagation [58].

### B. Pose-Only Visual-Reprojection Measurement

This part derives the residuals for pose-only VRM. From the pose-only multiply-view constraints (8), the landmark depth in the anchored keyframe $\mathbf{F}_{\varsigma}$ can be expressed as the function of the camera pose of the anchored keyframe $\mathbf{F}_{\varsigma}$ and $\mathbf{F}_{\eta}$ as

$$
\begin{aligned}
d_{\varsigma}^{(\varsigma,\eta)} &\triangleq \frac{\left\|\tilde{\boldsymbol{p}}^{u_{\eta}}\underset{\times}{}\boldsymbol{p}_{c_{\eta}c_{\varsigma}}^{c_{\eta}}\right\|}{\theta_{\varsigma,\eta}}, \\
\theta_{\varsigma,\eta} &= \left\|\tilde{\boldsymbol{p}}^{u_{\eta}}\underset{\times}{}\mathbf{R}_{c_{\varsigma}}^{c_{\eta}}\tilde{\boldsymbol{p}}^{u_{\varsigma}}\right\|,
\end{aligned}
\tag{19}
$$

where the relative pose $\left\{\boldsymbol{p}_{c_{\eta}c_{\varsigma}}^{c_{\eta}}, \mathbf{R}_{c_{\varsigma}}^{c_{\eta}}\right\}$ can be calculated using (3). Besides, the camera pose $\left\{\boldsymbol{p}_{wc}^{w}, \mathbf{R}_{c}^{w}\right\}$ can be written as the function of the IMU pose and the extrinsic parameters as

$$
\begin{cases}
\boldsymbol{p}_{wc}^{w} = \boldsymbol{p}_{wb}^{w} + \mathbf{R}_{b}^{w}\boldsymbol{p}_{bc}^{b}, \\
\mathbf{R}_{c}^{w} = \mathbf{R}_{b}^{w}\mathbf{R}_{c}^{b},
\end{cases}
\tag{20}
$$

where $\left\{\boldsymbol{p}_{wb}^{w}, \mathbf{R}_{b}^{w}\right\}$ is the IMU pose state in (16); $\left\{\boldsymbol{p}_{bc}^{b}, \mathbf{R}_{c}^{b}\right\}$ denotes the camera-IMU extrinsic parameters in (16).

For an observed keyframe $\mathbf{F}_{j}, j \neq \varsigma$ of the landmark $l_{i}$, the pose-only VRM residuals can be derived from (8) as

$$
\begin{aligned}
\boldsymbol{e}^{V}\left(\tilde{\boldsymbol{z}}_{j,l_{i}}^{V,\varsigma,\eta}, \boldsymbol{X}\right) &= \lfloor\boldsymbol{b}_{1}\quad \boldsymbol{b}_{2}\rfloor^{T}\left(\hat{\boldsymbol{p}}^{u_{j}} - \tilde{\boldsymbol{p}}^{u_{j}}\right), \\
\hat{\boldsymbol{p}}^{u_{j}} &= \frac{\hat{\boldsymbol{p}}^{c_{j}}}{\left\|\hat{\boldsymbol{p}}^{c_{j}}\right\|},
\end{aligned}
\tag{21}
$$

where $\hat{\boldsymbol{p}}^{u_{j}}$ is the calculated coordinate in the u-frame of the keyframe $\mathbf{F}_{j}$; $\boldsymbol{b}_{1} = \begin{bmatrix}1 & 0 & 0\end{bmatrix}^{T}$ and $\boldsymbol{b}_{2} = \begin{bmatrix}0 & 1 & 0\end{bmatrix}^{T}$ are two orthogonal bases [4], [34]. It should be noted that the residuals are equivalent to the two-view reprojection residuals if we have $\mathbf{F}_{j} = \mathbf{F}_{\eta}$. The covariance $\boldsymbol{\Sigma}_{l_{i}}^{V,\varsigma,\eta}$ is also propagated from the pixel plane onto the tangent plane, similar to (9). From (8), the calculated c-frame coordinate $\hat{\boldsymbol{p}}^{c_{j}}$ can be rewritten as

$$
\hat{\boldsymbol{p}}^{c_{j}} = d_{\varsigma}^{(\varsigma,\eta)}\mathbf{R}_{c_{\varsigma}}^{c_{j}}\tilde{\boldsymbol{p}}^{u_{\varsigma}} + \boldsymbol{p}_{c_{j}c_{\varsigma}}^{c_{j}},
\tag{22}
$$

By considering (19), $\boldsymbol{p}^{u_{j}}$ in (21) can be rewritten in more concise form

$$
\begin{aligned}
\hat{\boldsymbol{p}}^{u_{j}} &= \frac{\theta_{\varsigma,\eta}\hat{\boldsymbol{p}}^{c_{j}}}{\theta_{\varsigma,\eta}\left\|\hat{\boldsymbol{p}}^{c_{j}}\right\|} = \frac{\hat{\boldsymbol{p}}^{j}}{\left\|\hat{\boldsymbol{p}}^{j}\right\|}, \\
\hat{\boldsymbol{p}}^{j} &= \left\|\tilde{\boldsymbol{p}}^{u_{\eta}}\underset{\times}{}\mathbf{R}_{c_{\varsigma}}^{c_{\eta}}\tilde{\boldsymbol{p}}^{u_{\varsigma}}\right\|\left(\mathbf{R}_{c_{\varsigma}}^{c_{j}}\tilde{\boldsymbol{p}}^{u_{\varsigma}} + \boldsymbol{p}_{c_{j}c_{\varsigma}}^{c_{j}}\right),
\end{aligned}
\tag{23}
$$

where $\hat{\boldsymbol{p}}^{j}$ is a coordinate in a scaled camera frame (only used for better representation); the relative pose $\left\{\boldsymbol{p}_{c_{j}c_{\varsigma}}^{c_{j}}, \mathbf{R}_{c_{\varsigma}}^{c_{j}}\right\}$ can also be obtained from (3) and (20). Consequently, the calculated term $\hat{\boldsymbol{p}}^{u_{j}}$ is the function of the IMU poses $\left\{\boldsymbol{p}_{wb_{\varsigma}}^{w}, \mathbf{R}_{b_{\varsigma}}^{w}\right\}$, $\left\{\boldsymbol{p}_{wb_{j}}^{w}, \mathbf{R}_{b_{j}}^{w}\right\}$, and $\left\{\boldsymbol{p}_{wb_{\eta}}^{w}, \mathbf{R}_{b_{\eta}}^{w}\right\}$, and the camera-IMU extrinsic parameters $\left\{\boldsymbol{p}_{bc}^{b}, \mathbf{R}_{c}^{b}\right\}$. Consequently, the VRM residuals in (21) are in pose-only form, without involving the visual landmark states, as shown in Fig. 6.

### C. Pose-Only LiDAR-Depth Measurement

If a visual landmark is associated with a LiDAR depth, we should construct a pose-only LDM, as no depth state is available in the state vector. This part presents two pose-only LDM models, including the direct LDM and MSC-based LDM, and the latter is employed in PO-VINS. Note that the LiDAR depth $d_{i}$ associated with the landmark $l_{i}$ is always in the c-frame of the anchored keyframe $\mathbf{F}_{\varsigma}$, which is determined by the depth-association method [34].

#### 1) Direct LiDAR-Depth Measurement

As the landmark depth can be calculated using the IMU pose states and camera-IMU extrinsic parameters, a naive LDM model is to directly constrain the landmark depth expressed in (19). Hence, the residual for the direct LDM can be written as

$$
e^{D}\left(\tilde{\boldsymbol{z}}^{D_{i}}, \boldsymbol{X}\right) = \frac{\left\|\tilde{\boldsymbol{p}}^{u_{\eta}}\underset{\times}{}\boldsymbol{p}_{c_{\eta}c_{\varsigma}}^{c_{\eta}}\right\|}{\left\|\tilde{\boldsymbol{p}}^{u_{\eta}}\underset{\times}{}\mathbf{R}_{c_{\varsigma}}^{c_{\eta}}\tilde{\boldsymbol{p}}^{u_{\varsigma}}\right\|} - \tilde{d}_{i},
\tag{24}
$$

where $\tilde{d}_{i}$ is the LiDAR-depth observation for landmark $l_{i}$, whose two anchored keyframes are $\mathbf{F}_{\varsigma}$ and $\mathbf{F}_{\eta}$; the relative pose $\left\{\boldsymbol{p}_{c_{\eta}c_{\varsigma}}^{c_{\eta}}, \mathbf{R}_{c_{\varsigma}}^{c_{\eta}}\right\}$ is the same as that in (19). The LiDAR-depth residual in (24) is the function of the IMU poses $\left\{\boldsymbol{p}_{wb_{\varsigma}}^{w}, \mathbf{R}_{b_{\varsigma}}^{w}\right\}$ and $\left\{\boldsymbol{p}_{wb_{\eta}}^{w}, \mathbf{R}_{b_{\eta}}^{w}\right\}$, and the camera-IMU extrinsic parameters $\left\{\boldsymbol{p}_{bc}^{b}, \mathbf{R}_{c}^{b}\right\}$. Hence, the direct LDM is pose-only, which can directly constrain the pose states.

However, the VRM and LDM should be employed together for a single landmark to avoid a loss of observations. Thus, the measurement number may be increased inevitably, resulting in increased computational costs in solving the state-estimation problem. Besides, the LiDAR depth constraints in (24) are only imposed on the poses of the two anchored keyframes. As a consequence, the impacts of the LiDAR-depth outliers are lager, because not all pose states related to the landmark are employed in the direct LDM. All in all, the direct LDM may result in lower efficiency and reduced reliability.

*2) MSC-Based LiDAR-Depth Measurement*

In the famous MSCKF, the landmark states can be eliminated using the left nullspace projection, and a linearized measurement can be employed for the Kalman update [10], [11]. In short, the MSC update can be treated as a kind of pose-only form to some extent. Inspired by the MSC update, we propose an MSC-based LDM, which compresses all feature observations of a landmark and its LiDAR depth into a single measurement while retaining the pose-only form. As a result, the measurement number can be reduced because the pose-only VRM (presented in Section V.B) should not be used again to avoid information reuse. Besides, the impact of the LiDAR-depth outliers can also be reduced by employing all feature observations.

Considering a landmark $l_i$, its inverse-depth parameter in the anchored keyframe $\mathbf{F}_\zeta$ is expressed as $\tau_i$. For the feature observation in the keyframe $\mathbf{F}_j$, the residuals of the VRM in LE-VINS [34] can be written as

$$\boldsymbol{e}^V\left(\tilde{\boldsymbol{z}}_{j,l_i}^{V_c},\boldsymbol{X}\right) = \lceil \boldsymbol{b}_1 \quad \boldsymbol{b}_2 \rfloor^T \left| \frac{\hat{\boldsymbol{p}}^{c_j}}{\left\|\hat{\boldsymbol{p}}^{c_j}\right\|} - \tilde{\boldsymbol{p}}^{u_j} \right|, \tag{25}$$

where the calculated coordinate in the c-frame is expressed as

$$\hat{\boldsymbol{p}}^{c_j} = (\mathbf{R}_c^b)^T \left| (\mathbf{R}_{b_j}^w)^T \left| \mathbf{R}_{b_\zeta}^w \right| \frac{1}{\tau_i} \mathbf{R}_c^b \hat{\boldsymbol{p}}^{u_\zeta} + \boldsymbol{p}_{bc}^b \right| - \boldsymbol{p}_{bc}^b \right|, \tag{26}$$

where $\tilde{\boldsymbol{p}}^{u_j}$ and $\hat{\boldsymbol{p}}^{u_\zeta}$ denote the observed features in the u-frame. Similarly, the residual of the LDM in LE-VINS is calculated as

$$\boldsymbol{e}^D\left(\tilde{\boldsymbol{z}}^{D_i},\boldsymbol{X}\right) = \frac{1}{\tau_i} - \tilde{d}_i. \tag{27}$$

Finally, we obtain several VRMs from multiple observed keyframes and a LDM for the landmark $l_i$.

Linearizing the estimation errors for the states in (16) and the inverse-depth state $\tau_i$, the residuals in (25) and (27) can be approximated as

$$\boldsymbol{e}_i \simeq \mathbf{J}_{Y_i}\delta\boldsymbol{Y}_i + \mathbf{J}_{\tau_i}\delta\tau_i + \boldsymbol{n}_i, \tag{28}$$

where $\boldsymbol{Y}_i$ is a part of states related to the landmark $l_i$ in (16); $\boldsymbol{n}_i$ is the measurement noise matrix; $\mathbf{J}_{Y_i}$ and $\mathbf{J}_{\tau_i}$ are the corresponding Jacobians. We can now obtain a matrix $\mathbf{A}_i$, whose columns form a basis of the left nullspace of $\mathbf{J}_{\tau_i}$ [10], [11]. The residuals in (28) can be rewritten by left multiplying $\mathbf{A}_i^T$ on both sides as

$$\mathbf{A}_i^T\boldsymbol{e}_i \simeq \mathbf{A}_i^T\mathbf{J}_{Y_i}\delta\boldsymbol{Y}_i + \mathbf{A}_i^T\boldsymbol{n}_i. \tag{29}$$

Hence, we obtain the residuals of the MSC-based LDM as

$$\boldsymbol{e}_i^0 = \mathbf{A}_i^T\boldsymbol{e}_i \simeq \mathbf{J}_{Y_i}^0\delta\boldsymbol{Y}_i + \boldsymbol{n}_i^0, \tag{30}$$

where $\mathbf{J}_{Y_i}^0 = \mathbf{A}_i^T\mathbf{J}_{Y_i}$, and $\boldsymbol{n}_i^0 = \mathbf{A}_i^T\boldsymbol{n}_i$. The residuals $\boldsymbol{e}_i^0$ are now independent of the landmark inverse-depth errors and thus can be used for state estimation. As the dimensions of the matrix $\mathbf{J}_{\tau_i}$ for a single landmark is very small, we can derive $\mathbf{A}_i^T$ by employing a QR decomposition.

The MSC-based LDM residuals in (30) are expressed in pose-only form, without employing the landmark states, while incorporating both the visual-feature and the LiDAR-depth observations. Hence, we employ the pose-only MSC-based LDM for those landmarks with LiDAR depths; otherwise, the pose-only VDM is employed. Note that the proposed MSC-based LDM yields optimal state estimation, as the linearization errors can be reduced by multiple iterations during the optimization. In contrast, the conventional MSCKF suffers from one-time linearization, resulting in suboptimal estimation [16].

*D. Depth Estimation for Visual Landmarks*

The pose-only state estimator does not include the landmark depth states, yielding improved efficiency. The landmark depths can be analytically estimated to facilitate the outlier culling and the INS-aided feature tracking. For those landmarks without LiDAR depth, their depths are explicitly expressed by the pose of their anchored keyframes $\mathbf{F}_\zeta$ and $\mathbf{F}_{q_j}$. Hence, the landmark depth in the anchored keyframe $\mathbf{F}_\zeta$ can be updated by directly adopting the pose-only depth representation in (19) using the updated IMU pose and camera-IMU extrinsic parameters.

For those landmarks with LiDAR depths, we employ a depth-only optimization by combining the visual and LiDAR observations. Specifically, we use the VRM and LDM expressed in (25) and (27) to construct a new nonlinear optimizer. The IMU pose and camera-IMU extrinsic states are fixed without estimating, and only the inverse-depth states are estimated. The estimated inverse-depth states are employed to update the landmark depths and their positions in the w-frame. The LiDAR depth can be reflected to the landmark depth by using the depth-only optimization. Thus, the LiDAR-depth outliers can be further rejected in the outlier-culling module, which will be described in the next part. As a result, the state-estimation efficiency should be improved a little.

*E. Outlier Detection and Rejection*

Once the nonlinear optimization and depth estimation are finished, we conduct outlier detection and rejection for visual features. As the Huber robust cost function is employed in the estimator, the impacts of the outliers may not be notable during optimization. Nevertheless, they can be easily detected using error statistics. We can obtain the landmark positions in the w-frame with the estimated visual landmark depths. The reprojection errors can be calculated in each observed keyframe, and we can have a series of reprojection errors for the landmark.

If the average reprojection error exceeds the setting feature noise $\sigma_p$, this landmark, its feature observations, and the possible LiDAR depth will be treated as outliers. Otherwise, only those feature observations with a reprojection error larger than $3\sigma_p$ will be removed, while the visual landmark will be reserved. The reserved visual landmarks will be employed for the subsequent INS-aided feature tracking [4]. By adopting the outlier-culling method, those visual features affected by moving objects, repetitive textures, and illumination changes will be removed, and thus the system robustness can be improved.

## VI. EXPERIMENTS AND RESULTS

We conduct comprehensive experiments on public and private datasets with different carriers. We first introduce the datasets and the evaluation methods, and analyze the need for visual observations. Then, the accuracy and efficiency results are presented to examine the proposed PO-VINS. We also conduct a series of ablation experiments to evaluate the proposed methods fully. Finally, the real-time performance of PO-VINS on an onboard ARM computer is presented.

### A. Datasets and Evaluation Setup

#### 1) Datasets

The public and private datasets are adopted to evaluate the proposed methods fully. Besides, these datasets are collected by carriers with different dynamic conditions, including a low-speed robot, a handheld device, and a high-speed vehicle. Only a MEMS IMU, a camera, and a LiDAR are used on each dataset for quantitative evaluation. The details about the datasets are shown in Table I.

A low-speed wheeled robot is used in the private *Robot* dataset, and the maximum speed is about 1.5 m/s. The employed sensors in the *Robot* dataset include the MEMS-IMU ADIS16465, an RGB camera, and the solid-state LiDAR Mid-70 from Livox, as depicted in Fig. 7. The solid-state LiDAR Mid-70 has a non-repetitive and irregular scanning pattern, which is conducive for associating the visual features with LiDAR depths. The sensors are all well-synchronized through hardware triggers driven by a global navigation satellite system (GNSS) receiver. The high-accuracy ground-truth pose (0.02 m for position) is obtained by a post-processing real-time kinematic (RTK)/INS integrated navigation system [3]. Seven sequences, which are employed in LE-VINS [34], are used for a fair comparison. As the *Robot* dataset is collected on campus, the main challenging scenes are moving objects, illumination changes, and repetitive textures. Another two large-scale sequences are employed to evaluate the real-time performance of PO-VINS on the onboard ARM computer, i.e., the NVIDIA Xavier in Fig. 7.

The adopted public datasets are the *R3LIVE* [48] and *FusionPortableV2* [61]. The *R3LIVE* dataset includes indoor and outdoor environments, making it highly challenging. Besides, large angular motions may frequently happen due to the handheld carrier, and thus, motion blurs occur, especially in dim indoor environments. As no ground truth is included in the

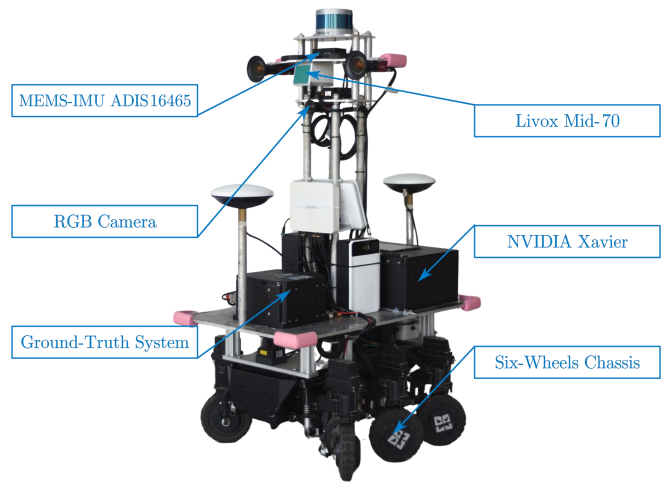

Fig. 7. The equipment setup in the *Robot* dataset, and the carrier is a low-speed wheeled robot. All sensors are well-synchronized though hardware trigger, and the dataset is collected by the on-board ARM computer (Xavier).

TABLE I
DATASETS DESCRIPTIONS

| Datasets | *Robot* | *R3LIVE* | *FusionPortableV2* |
|---|---|---|---|
| Carrier | Wheeled robot | Handheld | Vehicle |
| Camera | 1280*1024, 20 Hz | 1280*720, 15 Hz | 1024*768, 20 Hz |
| IMU | ADI ADIS16465, 200Hz | BOSCH BMI088, 200 Hz | Sensonor STIM300, 200Hz |
| LiDAR | Livox Mid-70, 10 Hz | Livox AVIA, 10 Hz | Ouster OS1-128, 10 Hz |
| Length | 7 sequences with 9550 s and 13.4 km | 6 sequences with 4462 s and 5.0 km | 5 sequences with 2527 s and 20.8 km |

*R3LIVE* dataset, the end-to-end errors can be utilized for quantitative evaluation. Hence, only six sequences with end-to-end trajectories are adopted in the *R3LIVE* dataset. For the *FusionPortableV2* dataset, the five vehicle sequences with good ground truth are used. As the vehicle travels very fast, the *FusionPortableV2* dataset is more challenging than the other two datasets, especially for the *highway* sequences. The RTK/INS integrated navigation results are employed as the ground truth in the *FusionPortableV2* dataset. More details about the two datasets can be found in Table I.

#### 2) Evaluation Setup

The proposed PO-VINS is implemented by building upon LE-VINS [34], which is available online. Hence, LE-VINS is employed as the baseline system for PO-VINS. The pure VINS systems without the LiDAR enhancement are expressed as LE-VINS-WO and PO-VINS-WO, respectively. As PO-VINS is an optimization-based system, the state-of-the-art VINS VINS-Mono [19] is employed for comparisons. Besides, the tightly-coupled LiDAR-visual-inertial navigation systems (LVINS) LVI-SAM [51] and R2LIVE [62] are also involved. Note that they are all feature-based systems and direct-based systems are not engaged for fair comparisons. We extract 150 visual features for these systems on the *Robot* dataset due to its lower dynamic, while 200 features on the *R3LIVE* and *FusionPortableV2* datasets. According to our experiments, the sliding-window size is set to 10 for LE-VINS-WO, LE-VINS,

PO-VINS-WO, and PO-VINS to bound the accuracy and efficiency. These systems are all run in real time within the robot operation system (ROS) framework on a laptop (Intel i7-13700H and NVIDIA RTX3050). Note that no explicit loop closure is used for these systems.

As the ground-truth pose is included in the *Robot* and *FusionPortableV2* datasets, the absolute translation error (ATE) [63] is utilized for quantitative evaluations. Specifically, the EVO tool [64] is adopted to calculate the ATEs. Besides, we also employ the relative translation error (RTE) [63] to evaluate the robustness of PO-VINS. The reason is that the RTE over a distance such as 25 m, can reflect the short-time consistence. For the *R3LIVE* dataset, the 3D end-to-end errors are used for evaluations.

### B. Analysis of the Need for Visual Observations

In PIPO-SLAM [31], the configuration of 1000 cameras and 100000 landmarks is employed to evaluate the performance improvement of the PA in the local-mapping thread. This configuration is not suitable for real-time applications, as it is very slow even when using the PA (about 68 seconds). Besides, the benefits and costs of employing such a large-scale state-estimation problem are unbalanced. Hence, we analyze the need for visual observations to bound the efficiency and accuracy by detecting different features and using different sliding-window sizes.

Fig. 8 shows ATEs and time costs of the state estimation by LE-VINS-WO over different configurations on the low-speed *Robot-Exp-6* and high-speed *FusionPortableV2-highway01* sequences. Here, LE-VINS-WO is adopted as it is a pure VINS without being affected by LiDAR-related errors. According to the results, detecting more visual features may result in smaller ATE while lower state-estimation efficiency to some extent. However, we can only extract limited high-quality visual features from the image for optical-flow tracking. Hence, detecting too many features is somewhat meaningless, as the accuracy improvement is very small while the computational costs are increased. Hence, we only extract 150~200 visual features according to the carrier dynamic. The sliding-window size has a similar impact on the localization accuracy. However, when the sliding-window size is larger than 30, the VINS system yields a little degraded accuracy, while the time costs increase notably. There are many reasons for the degeneration, such as wrongly removing visual observations by outlier-culling algorithm and inaccurately estimated IMU biases. Nevertheless, the results are very normal for the complex VINS systems.

The results in Fig. 8 demonstrate that detecting too many visual features or using a too-large sliding window may not always result in improved positioning accuracy, while may increase the time costs. Nevertheless, it is meaningful to bound the accuracy and efficiency by setting proper hyper-parameters, i.e., feature number and sliding-window size. In other words, the state-estimation dimensions in a real-time VINS should not be too large. Suppose that the VINS may include only ten visual keyframes and 100~200 landmarks in the sliding window. As the inverse-depth parameter is employed in LE-VINS-WO, the

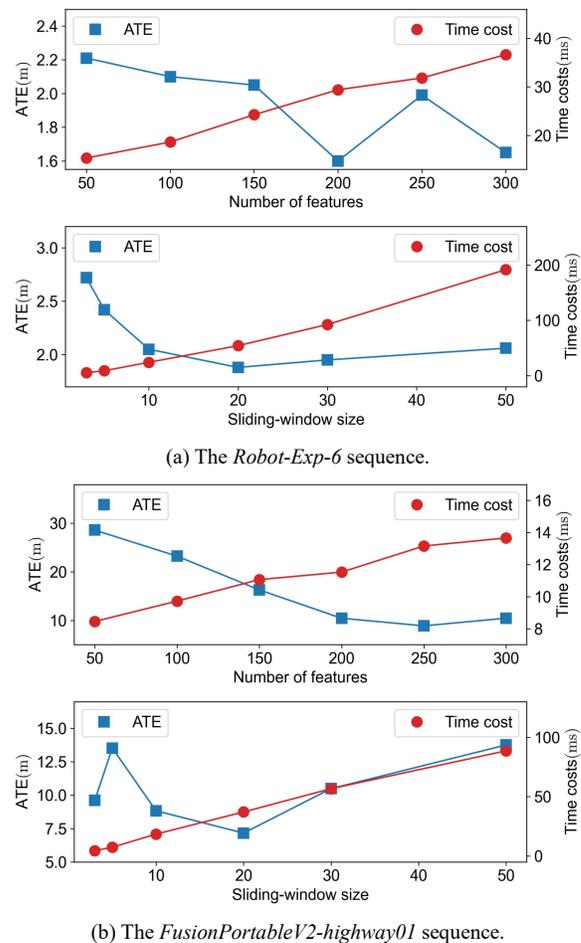

(a) The *Robot-Exp-6* sequence.

(b) The *FusionPortableV2-highway01* sequence.

Fig. 8. The ATE and the time costs of the state estimation by LE-VINS-WO (without LiDAR enhancement) over different configurations. The number of features denotes the maximum visual features detected in the image.

landmark states is about 100~200 dimensions. As an IMU state is 16 dimensions according to (16), the landmark states almost yield the same dimensions as the total IMU states in the sliding window. Consequently, the efficiency improvement by using the pose-only visual representation should not be larger than 100% in the real-time and tightly-coupled VINS.

### C. Evaluation of the Accuracy

#### 1) Private Robot Dataset

The evaluation results on the private *Robot* dataset is shown in Table II. The seven sequences are originally employed in LE-VINS, and they are all collected in complex campus environments. According to the results, LE-VINS and PO-VINS yield notably improved accuracy than SOTA methods VINS-Mono, LVI-SAM, and R2LIVE regarding the average ATE. This benefits from the INS-centric design, and thus the INS accuracy can be performed entirely [4]. Nevertheless, LVI-SAM and R2LIVE exhibit the best results on *Exp-1*. As *Exp-1* is in a small-scale environment, the LiDAR subsystem in them can match with their built point-cloud map, and thus the drifts can be eliminated. We further show the position errors for LVINSs on *Exp-5*, as depicted in Fig. 9. LE-VINS and PO-VINS exhibit smaller drifts along the z-axis than LVI-SAM and R2LIVE. PO-VINS also exhibits superior

TABLE II
COMPARISON OF THE ATEs ON THE *ROBOT* DATASET

| ATE (m) | VINS-Mono | LVI-SAM | R2LIVE | LE-VINS-WO | LE-VINS | PO-VINS-WO | PO-VINS |
|---------|-----------|---------|--------|------------|---------|------------|---------|
| *Exp-1* | 4.67 | 9.16 | 5.05 | 1.86 | 1.46 | 1.74 | 1.5 |
| *Exp-2* | 2.53 | 1.96 | 2.62 | 1.28 | 1.16 | 1.52 | 1.2 |
| *Exp-3* | 2.51 | 2.05 | 1.14 | 0.84 | 0.81 | 0.86 | 0.77 |
| *Exp-4* | 1.65 | 0.82 | 0.67 | 0.93 | 0.84 | 1.05 | 0.9 |
| *Exp-5* | 3.64 | 7.37 | 2.48 | 1.53 | 2.45 | 1.71 | 0.8 |
| *Exp-6* | 5.39 | 5.19 | 2.02 | 2.05 | 1.61 | 1.78 | 1.86 |
| *Exp-7* | 3.73 | 2.24 | 3.06 | 1.34 | 0.76 | 0.75 | 0.81 |
| **Average** | **3.45** | **4.11** | **2.43** | **1.40** | **1.30** | **1.34** | **1.12** |

LE-VINS-WO and PO-VINS-WO denote the configurations without LiDAR enhancement. The red result represents the best among different methods, while the green result is the second best.

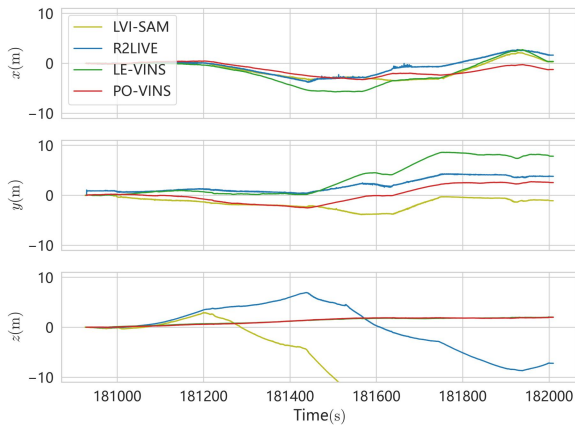

Fig. 9. Position errors on the *Robot-Exp-5* sequence with the initial point aligned. Here, only the LVINSs are employed for better visualization.

performance along the x-axis and y-axis, while LE-VINS shows the largest drift along the y-axis.

Besides, PO-VINS-WO almost achieves the same average accuracy as LE-VINS-WO in Table II. It demonstrates that the pose-only visual representation is suitable for tightly-coupled VINS without losing accuracy. Compared to LE-VINS-WO, LE-VINS also exhibits improved accuracy by incorporating the LiDAR enhancement on all sequences except for *Exp-5*. The reason for the degradation on *Exp-5* may be the LiDAR-depth outliers. In contrast, PO-VINS yields the best on *Exp-5*, and the proposed uncertainty-based outlier rejection is one of the reasons. The main reason for the improvement is the employed MSC-based LDM, because PO-VINS is more robust for LiDAR-depth outliers by employing all visual features and LiDAR depth to construct the MSC-based LDM. More ablation experimental results are presented in Section VI.E. The average absolute translation accuracy of PO-VINS is improved by 16.4% compared to PO-VINS-WO. As the original VINS system PO-VINS-WO already achieves satisfactory accuracy, the improvement for PO-VINS is not very significant. Note that the minor degradation on *Exp-6* and *Exp-7* for PO-VINS should be a normal phenomenon, as the system is affected by many factors, especially in such complex experimental environments.

As the short-term relative errors can reflect the robustness to

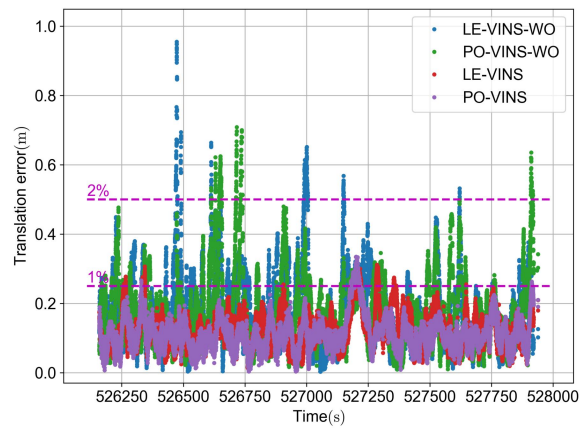

(a) The *Robot-Exp-2* sequence.

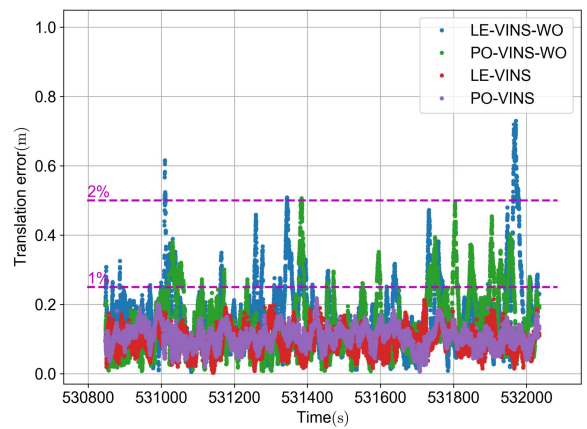

(b) The *Robot-Exp-7* sequence.

Fig. 10. Comparison of the RTEs over 25 m.

some extent, we also evaluate RTEs over 25 m to evaluate the proposed pose-only methods. Note that the rotation or attitude errors can be reflected in the RTE; thus, only RTE is employed [63]. The RTEs over 25 m on *Exp-2* and *Exp-7* are shown in Fig. 10. According to the results, LE-VINS-WO and PO-VINS-WO almost achieve a similar RTE, though the maximum RTE for PO-VINS-WO on each sequence is much smaller. Hence, the results illustrate that the proposed pose-only VINS can achieve similar robustness regarding short-time accuracy. With the

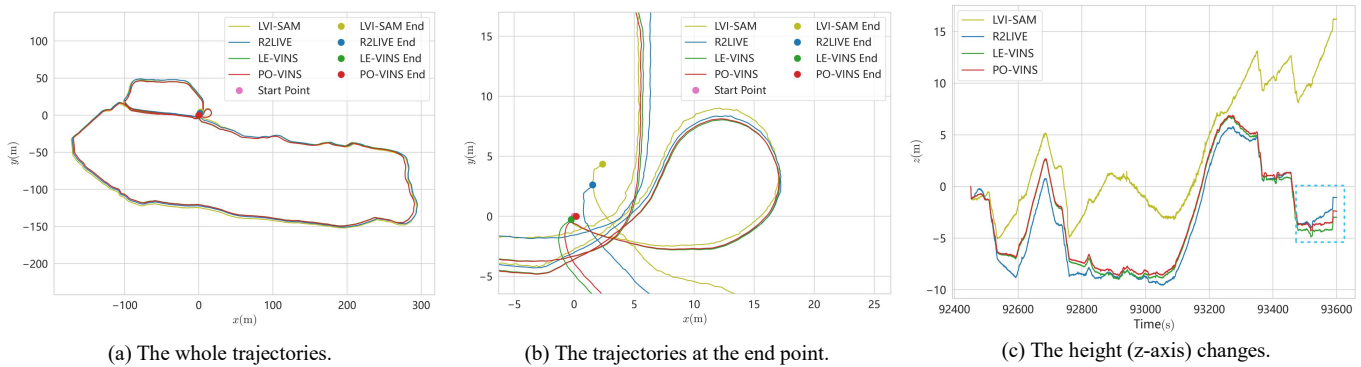

(a) The whole trajectories.  (b) The trajectories at the end point.  (c) The height (z-axis) changes.

Fig. 11. The trajectories on the *R3LIVE-hkust-campus01* sequence for LVINSs. The cyan rectangle in (c) denotes the abnormal height result for R2LIVE.

TABLE III
COMPARISON OF THE END-TO-END ERRORS ON THE *R3LIVE* DATASET

| ATE (m) | VINS-Mono | LVI-SAM | R2LIVE | LE-VINS-WO | LE-VINS | PO-VINS-WO | PO-VINS |
|---------|-----------|---------|--------|-----------|---------|-----------|---------|
| *hku-main-building* | failed | 8.33 | 0.54 | 2.70 | 0.46 | 4.03 | 0.33 |
| *hkust-campus00* | failed | 15.69 | 0.05 | 13.76 | 12.59 | 11.43 | 10.72 |
| *hkust-campus01* | failed | 17.00 | 3.28 | 5.71 | 3.03 | 5.72 | 2.43 |
| *hkust-campus02* | 11.67 | 15.05 | 1.63 | 5.61 | 3.06 | 6.14 | 2.54 |
| *hku-park0* | 1.60 | 2.59 | 0.08 | 0.76 | 1.36 | 1.39 | 1.64 |
| *hku-park1* | 26.61 | 0.55 | 0.55 | 1.11 | 0.84 | 1.27 | 0.83 |
| **Average** | **Invalid** | **9.87** | **1.02** | **4.94** | **3.55** | **5.00** | **3.08** |

LiDAR enhancement, PO-VINS also exhibits the same RTE as LE-VINS. It demonstrates that the proposed pose-only MSC-based LDM can perform the same local consistency as the conventional LiDAR-enhanced VINS.

*2) Public R3LIVE-Handheld Dataset*

The low-speed *Robot* dataset is less challenging in terms of dynamic conditions. Hence, the *R3LIVE-Handheld* dataset is adopted to evaluate the proposed methods, as large angular motions may frequently happen. The end-to-end errors for the employed systems are shown in Table III, and Fig. 11 depicts the trajectories on the *hkust-campus01* sequence for LVINSs. According to the results, VINS-Mono fails on three sequences, manly because of the motion blur caused by large angular motion in indoor environments. Besides, the visual-inertial time-delay parameter has changed notably over time. Thus, VINS-Mono can not estimate the changed time-delay parameter, which is modeled as a random constant. In contrast, LE-VINS-WO and PO-VINS-WO succeed in running on all the sequences benefiting from the INS-centric design. Meanwhile, we model the visual-inertial time-delay parameter as a random walk, and thus, it can be estimated accurately over time. R2LIVE yields the best average results on the *R3LIVE-Handheld* dataset, as the LiDAR subsystem can perform better in such environments. LVI-SAM exhibits worse results because it is designed for spinning LiDARs rather than the Livox LiDAR employed in the *R3LIVE* dataset. Nevertheless, R2LIVE can match its previously built point-cloud map, resulting in lower end-to-end errors, according to the results in Fig.11.c. That is why R2LIVE can achieve centimeter-level end-to-end errors on some sequences. In other words, the R2LIVE may exhibit worse results, if the

ATE can be used for evaluation.

The pure VINSs PO-VINS-WO and LE-VINS-WO show similar average errors. With the LiDAR-depth enhancement, LE-VINS and PO-VINS yield notably improved accuracy than their VINS systems on all the sequences except for *hku-park0*. The reason is that *hku-park0* is in a dim park with many unstructured trees and bushes. Thus, only a few valid visual features will be reserved after using the outlier-rejection method (Section V.E) due to the impacts of the LiDAR-depth outliers. Nevertheless, PO-VINS shows a lower average error than LE-VINS. Besides, the proposed PO-VINS achieves the second-best accuracy regarding the average error and the best accuracy on two sequences. Overall, LE-VINS demonstrates superior robustness than the baseline system LE-VINS on the *R3LIVE* dataset and even exhibits higher accuracy than tightly-coupled LVIO R2LIVE on some sequences.

*3) Public FusionPortableV2-Vehicle Dataset*

As the carrier is very slow on the previous datasets, about several meters per second, the high-speed *FusionPortableV2-Vehicle* dataset is employed. However, we fail to run LVI-SAM and R2LIVE because of sensor incompatibility. The ATEs on the *FusionPortableV2* dataset are shown in Table IV. VINS-Mono exhibits the worst ATE, especially on high-speed highway sequences. PO-VINS-WO shows a lower average ATE than LE-VINS-WO, which is caused by the notable difference on the *highway00* sequence. Nevertheless, LE-VINS-WO and PO-VINS-WO almost yield similar accuracy on other sequences. PO-VINS shows improved accuracy than PO-VINS-WO by incorporating the LiDAR enhancement, especially on the highway sequences. PO-VINS also exhibits superior accuracy to LE-VINS,

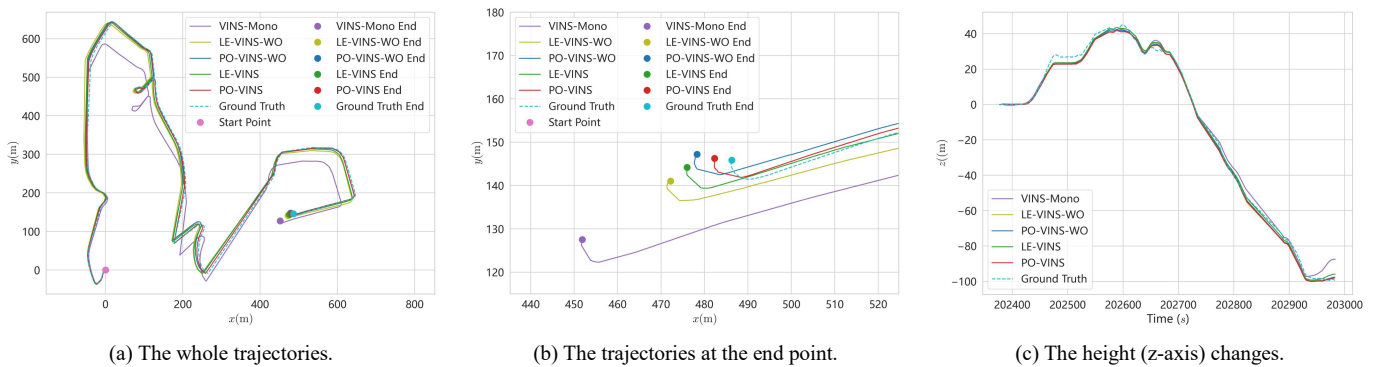

(a) The whole trajectories.

(b) The trajectories at the end point.

(c) The height (z-axis) changes.

Fig. 12. The trajectories on the *FusionPortableV2-campus00* sequence.

<table>
<tr><td colspan="6" align="center">TABLE IV<br>COMPARISON OF THE ATEs ON THE *FUSIONPORTABLEV2* DATASET</td></tr>
</table>

| ATE (m) | VINS-Mono | LE-VINS-WO | LE-VINS | PO-VINS-WO | PO-VINS |
|---|---|---|---|---|---|
| *campus00* | 24.37 | 6.32 | 4.49 | 3.75 | 2.48 |
| *campus01* | 13.87 | 6.35 | 5.99 | 6.22 | 6.90 |
| *downhill00* | 71.6 | 4.05 | 3.43 | 4.58 | 2.81 |
| *highway00* | failed | 31.98 | 25.91 | 13.88 | 9.27 |
| *highway01* | 141.98 | 12.9 | 7.99 | 10.45 | 7.93 |
| **Average** | **Invalid** | **12.32** | **9.56** | **7.78** | **5.88** |

TABLE V
COMPARISON OF THE STATE-ESTIMATION TIME ON THE *ROBOT* DATASET

| Time (ms) | LE-VINS-WO | LE-VINS | PO-VINS-WO | PO-VINS |
|---|---|---|---|---|
| *Exp-1* | 24.6 | 24.5 | 16.2 | 16.3 |
| *Exp-2* | 22.8 | 24.2 | 15.6 | 16.2 |
| *Exp-3* | 22.0 | 24.3 | 13.8 | 16.1 |
| *Exp-4* | 22.0 | 24.7 | 14.3 | 15.9 |
| *Exp-5* | 24.5 | 24.6 | 15.7 | 16.2 |
| *Exp-6* | 24.6 | 24.2 | 15.5 | 16.7 |
| *Exp-7* | 21.0 | 23.2 | 15.7 | 16.2 |
| **Average** | **23.1** | **24.2** | **15.3** | **16.2** |

The time costs for the state estimation include the factor graph optimization and the marginalization.

benefiting from the proposed pose-only solutions for LiDAR depth. Fig. 12 depicts the trajectories on the *campus00* sequence. PO-VINS shows a better-aligned trajectory to ground truth than other systems, and the end point is the closest to the ground truth. In conclusion, the results demonstrate that the proposed pose-only methods are robust on the high-speed vehicle dataset and achieve improved accuracy compared to the baseline system LE-VINS.

### D. Evaluation of the Efficiency

The accuracy results in the previous section illustrate that the pose-only VINS, i.e., PO-VINS-WO, exhibits a similar accuracy to the conventional baseline system LE-VINS-WO. Besides, the pose-only LiDAR-enhanced VINS, i.e., PO-VINS, yields improved accuracy and robustness to LE-VINS by incorporating the proposed uncertainty-based outlier rejection and MSC-based LDM. Nevertheless, the main advantage of the pose-only solution lies in the higher state-estimation efficiency. The statistical results of the state-estimation time, which includes the factor graph optimization and the marginalization, are shown in Table V.

Compared to the VINS baseline LE-VINS-WO, the average state-estimation time of PO-VINS-WO is reduced by about 33.8%. According to the analysis in Section VI.B, the results are reasonable, as the dimension of the state vector can only be reduced by about 50% by using the pose-only VRM. The time costs for LE-VINS increased a little as extra LiDAR-depth measurements are employed. PO-VINS exhibits shorter time costs, and the average state-estimation time is reduced by 33.1% compared to the baseline LE-VINS. This benefits from the proposed MSC-based LDM and the depth-only optimization. The MSC-based LDM can combine all visual features and the LiDAR depth observations of a visual landmark into a single measurement, and thus the measurement number can be reduced notably. Meanwhile, the LiDAR-depth outliers can be exposed by using depth-only optimization, resulting in fewer measurements while maintaining superior robustness. Ablation experiments will be conducted in the next section to evaluate the impacts of these employed methods.

The efficiency results demonstrate that the pose-only solution can also achieve notable improved state-estimation efficiency in the tightly-coupled VINS, even with LiDAR enhancement. Besides, the efficiency improvement should be more significant on the onboard ARM computer with limited computational resources. Although the improvement is not larger than 100% in such a real-time configuration due to the small dimension of the state-estimation problem, the results are satisfied to improve the real-time performance.

### E. Ablation Experiments

Ablation experiments are conducted to evaluation the impacts of the uncertainty-based outlier-rejection method, MSC-based LDM, and depth-only optimization. Table VI exhibits the ATEs and state-estimation time by using different configurations.

#### 1) The Impact of the Uncertainty-Based Outlier Rejection

The proposed uncertainty-based outlier-rejection method can detect and reject the LiDAR-depth outliers and thus can improve the localization accuracy. As shown in Table VI, the ATEs are increased on four sequences without using the proposed outlier-rejection method, especially on the *Exp-1* and

TABLE VI
THE ATEs AND THE STATE-ESTIMATION TIME ON THE *Robot* DATASET WITH
DIFFERENT CONFIGURATIONS

| | w/o outlier rejection[1] | | Using direct LDM[2] | | w/o depth-only optimization[3] | | Proposed PO-VINS | |
|---|---|---|---|---|---|---|---|---|
| | ATE (m) | Time (ms) | ATE (m) | Time (ms) | ATE (m) | Time (ms) | ATE (m) | Time (ms) |
| *Exp-1* | 2.27 | 16.7 | 2.2 | 22.1 | 1.74 | 21.3 | 1.50 | 16.3 |
| *Exp-2* | 1.18 | 16.2 | 1.69 | 21.9 | 1.38 | 19.8 | 1.20 | 16.2 |
| *Exp-3* | 0.77 | 16.6 | 0.66 | 19.9 | 0.81 | 19.3 | 0.77 | 16.1 |
| *Exp-4* | 0.85 | 16.3 | 1.25 | 20.0 | 0.86 | 18.8 | 0.90 | 15.9 |
| *Exp-5* | 1.01 | 17.6 | 1.49 | 21.0 | 0.57 | 20.7 | 0.80 | 16.2 |
| *Exp-6* | 2.89 | 16.7 | 2.37 | 20.4 | 1.70 | 20.6 | 1.86 | 16.7 |
| *Exp-7* | 1.02 | 15.6 | 1.01 | 20.9 | 0.85 | 18.4 | 0.81 | 16.2 |
| **Average** | **1.43** | **16.5** | **1.52** | **20.9** | **1.13** | **19.8** | **1.12** | **16.2** |

[1]Without using the uncertainty-based outlier rejection method proposed in Section IV.C.
[2]Using the direct LDM in Section V.C.1.
[3]Without using the depth-only optimization presented in Section V.D.

*Exp-6* sequences. Meanwhile, the ATEs are almost the same on the other three sequences. The average ATE of the proposed PO-VINS is reduced by about 21.7% compared to the configuration without using the outlier rejection. Besides, the state-estimation time costs are almost the same with or without the outlier rejection. The state-estimation efficiency is improved a little for PO-VINS, as the LiDAR-depth measurements are decreased after the outlier rejection. The ablation results demonstrate that the uncertainty-based outlier rejection for LiDAR depth can effectively detect and reject outliers and improve localization accuracy. Note that the proposed outlier rejection can run fast, and the average time cost is about 0.2 ms.

*2) The Impact of the MSC-Based LiDAR Depth Measurement*

The proposed pose-only MSC-based LDM combines all the visual feature and LiDAR depth observations of a landmark into a single measurement. Hence, it can improve both the robustness and the efficiency, as illustrated in Table VI. Compared to the direct LDM configuration, the proposed PO-VINS is more robust to LiDAR-depth outliers, as all feature observations are employed. In contrast, only the feature observations in the two anchored keyframes are used in the direct LDM. Meanwhile, the LDM number can be decreased notably by combining all related observations into one measurement, and thus the state-estimation efficiency can also be reduced. The results exhibit that the proposed pose-only MSC-based LDM is accurate and efficient.

*3) The Impact of the Depth-Only Optimization*

The depth-only optimization is conducted to update the depths of those landmarks associated with the LiDAR depth. According to the results in Table VI, the proposed PO-VINS yield improved efficiency to the configuration without the depth-only optimization while achieving similar ATEs. The efficiency improvement is because the LiDAR-depth outliers can be exposed after the depth-only optimization and can be detected and rejected. As a result, the LDM number can be decreased a little, and state-estimation efficiency can be improved without sacrificing the accuracy. The average time cost of the depth-only optimization is about 1.8 ms. Thus, it is meaningful to employ the proposed depth-only optimization.

*F. Real-Time Performance*

We also carry on experiments on an onboard ARM computer (NVIDIA Xavier) to evaluate the real-time performance of the proposed PO-VINS. The trajectories of the employed large-scale sequences *Seq-1* (2012 s and 2565 m) and *Seq-2* (1162 s and 1462 m) are depicted in Fig. 13. The two sequences are also collected by the wheeled robot depicted in Fig. 7. The same configurations on the previous experiments, which are running on the laptop, are adopted on the onboard computer for LE-VINS and PO-VINS. As shown in Table VII, the front-end time for PO-VINS is almost the same, while the average back-end time is reduced by 56.3%, compared to the baseline system LE-VINS. Note that both PO-VINS and LE-VINS can run in real time on the onboard computer, even with the same configurations employed on the laptop. This is because the FGO is only conducted when a new keyframe is selected, and the average keyframe interval is about 0.3 s. It demonstrates that the proposed pose-only solution is more efficient on the onboard ARM computer with limited computational resources. Meanwhile, PO-VINS yields notably improved accuracy to LE-VINS on the two sequences regarding the ATE.

We also evaluate the total running time of LE-VINS and PO-VINS and calculate the equivalent frame per second (FPS), as shown in Table VIII. Here, the equivalent FPS is calculated by dividing the sequence length by the running time and multiplying by the camera frame rate (20 FPS). The total running time of PO-VINS on the two sequences are decreased by 28.2% and 26.5%, respectively. Besides, the FPSs of

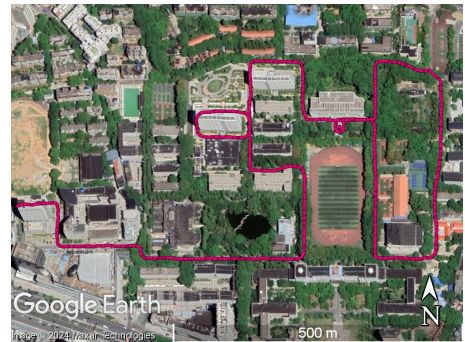

(a) The test trajectory on the *Seq-1* (2012 s and 2565 m).

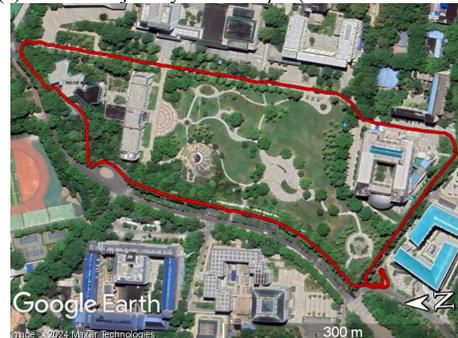

(b) The test trajectory on the *Seq-2* (1162 s and 1462 m).
Fig. 13. The trajectories on the real-time experiments.

TABLE VII
THE PROCESSING TIME AND ATES ON THE REAL-TIME EXPERIMENTS

| | Front-end[1] (ms) | | Back-end[2] (ms) | | ATE (m) | |
|---|---|---|---|---|---|---|
| | LE-VINS | PO-VINS | LE-VINS | PO-VINS | LE-VINS | PO-VINS |
| *Seq-1* | 28 | 26 | 109 | 48 | 2.82 | 2.05 |
| *Seq-2* | 25 | 24 | 104 | 45 | 2.20 | 1.64 |
| **Average** | **26.5** | **25.0** | **106.5** | **46.5** | **2.51** | **1.85** |

[1]The front-end processing time includes image preprocessing, feature detection and tracking, and the depth association.
[2]The back-end processing time, *i.e.*, the state-estimation time, includes the factor graph optimization and the marginalization.
Note that the depth association and the factor graph optimization are only conducted when a new keyframe is selected, and the average keyframe interval is about 0.3 s. Thus, both LE-VINS and PO-VINS can run real-time on the on-board ARM computer.

TABLE VIII
THE TOTAL RUNNING TIME AND EQUIVALENT FPS ON THE REAL-TIME EXPERIMENTS

| | Sequence length (s) | Total running time (s) | | Equivalent FPS | |
|---|---|---|---|---|---|
| | | LE-VINS | PO-VINS | LE-VINS | PO-VINS |
| *Seq-1* | 2012 | 1315 | **944** | 30.6 | **42.6** |
| *Seq-2* | 1162 | 830 | **610** | 28.0 | **38.1** |

The equivalent FPS is calculated by dividing the sequence length by the running time and multiplying by the camera frame rate, and the camera frame rate is 20 FPS.

PO-VINS are increased by 39.2% and 36.1%, respectively. The real-time experimental results demonstrate that the proposed pose-only solution and PO-VINS effectively improve the estimation efficiency on the onboard ARM computer. Hence, PO-VINS can be a high-efficiency and robust solution for VISE with an optional LiDAR enhancement.

## VII. CONCLUSIONS AND DISCUSSIONS

This study propose a tightly-coupled LiDAR-enhanced VINS using a full pose-only solution to achieve an efficient and robust state estimation. In the proposed pose-only VINS, the landmark depths are explicitly expressed by the pose of the anchored keyframe. Hence, the landmark states can be explicitly removed from the state vector. The pose-only VINS yields a similar accuracy to the conventional VINS, while the state-efficiency is improved by more than 30% even in the real-time configuration. Hence, the results demonstrate that the pose-only visual representation is applicable for real-time VINS. The analytical depth uncertainty for visual landmarks can also be derived within the pose-only framework. Hence, we can employ the depth uncertainty to detect and reject LiDAR depth outliers, which has been proven to be effective. Besides, we propose an MSC-based LDM to seamlessly incorporate the LiDAR enhancement into the pose-only solution. The proposed PO-VINS achieves improved robustness and efficiency compared to the baseline system LE-VINS on several datasets with different carriers. Specifically, the state-estimation efficiency is improved by more than 30% and 50% on the laptop and the onboard ARM computer, respectively.

We believe that the pose-only VISE can be a substitute for the conventional BA-based methods, as it is more efficient while maintaining the same accuracy. Future works will incorporate LiDAR odometry to construct a tightly-coupled LVINS, while the LiDAR-depth enhancement can also be employed. We also intend to explore further applications of the pose-only visual representation.